\documentclass[chaparabic]{clv3}
\usepackage{alocal}

\usepackage[T1]{fontenc}
\usepackage[utf8]{inputenc}
\usepackage{booktabs}
\usepackage{tikz}
\usepackage{pbox}
\usepackage{mathtools}
\usepackage{wrapfig,lipsum,booktabs}
\usepackage{graphics}
\usepackage{lastpage}
\usepackage{pdfpages}
\usepackage[hidelinks]{hyperref}

\usepackage[flushleft]{threeparttable}
\usepackage{array}

\usepackage{xcolor}

\newcommand{\proposedwork}{Segmentation Methods and Neural Models for Sentiment Analysis}

\newcommand{\datasetproducts}{\emph{Product Reviews}}
\newcommand{\datasetbeyazperde}{\emph{Movie Reviews}}

\newcommand{\modelchar}{\emph{Character}}
\newcommand{\modellemma}{\emph{Lemma}}
\newcommand{\modelstem}{\emph{Stem}}
\newcommand{\modelstemsuffix}{\emph{Stem+Suffix}}
\newcommand{\modelstemmeta}{\emph{Stem+Suffix/Meta}}
\newcommand{\modellemmasuffix}{\emph{Lemma+Suffix}}
\newcommand{\modellemmameta}{\emph{Lemma+Suffix/Meta}}
\newcommand{\modelbpmini}{\emph{BPE-1k}}
\newcommand{\modelbpmid}{\emph{BPE-5k}}
\newcommand{\modelbplarge}{\emph{BPE-30k}}
\newcommand{\modelhybrid}{\emph{Hybrid}}
\newcommand{\modeltoken}{\emph{Word-Token}}
\newcommand{\modeltokenmeta}{\emph{Token-Meta}}
\newcommand{\modelsyllable}{\emph{Syllable}}

\newcommand{\baseword}{\emph{Word-based}}
\newcommand{\basesubword}{\emph{Sub-word}}
\newcommand{\basemorph}{\emph{Morphology-based}}
\newcommand{\basehybrid}{\emph{Hybrid}}
\newcommand{\basemodels}{\baseword{}, \basesubword{}, \basemorph{}, and \basehybrid{}}
\newcommand{\basemodelscount}{\emph{4}}

\newcommand{\wordbasedlist}{\modeltoken{}}
\newcommand{\bpelist}{\modelbpmini{}, \modelbpmid{}, \modelbplarge{}}
\newcommand{\subwordlist}{\modelchar{}, \bpelist{}, \modelsyllable{}}
\newcommand{\morphlist}{\modellemma{}, \modellemmasuffix{}, \modellemmameta{}, \modelstem{}, \modelstemsuffix{}, \modelstemmeta{}, \modeltokenmeta{}}
\newcommand{\hybridlist}{\modelhybrid{}}

\issue{}{}{}

\dochead{}

\runningtitle{Effects of Segmentation Models for Neural Model based Sentiment Analysis}

\runningauthor{F. Kurt}

\begin{document}
	
\title{Investigating the Effect of Segmentation Methods on Neural Model based Sentiment Analysis on Informal Short Texts in Turkish}

\author{Fatih Kurt \thanks{Informatics Institute, Ankara, Turkey}}
\affil{Middle East Technical University (METU)} 

\author{Dilek Kisa \thanks{Computer Eng. Dept., Ankara, Turkey}}
\affil{Middle East Technical University (METU)}

\author{Pinar Karagoz \thanks{Computer Eng. Dept., Ankara, Turkey}}
\affil{Middle East Technical University (METU)}
	
\maketitle

\begin{abstract}
This work investigates segmentation approaches for sentiment analysis on informal short texts in Turkish. The two building blocks of the proposed work are segmentation and deep neural network model. Segmentation focuses on preprocessing of text with different methods. These methods are grouped in four: morphological, sub-word, tokenization, and hybrid approaches. We analyzed several variants for each of these four methods. The second stage focuses on evaluation of the neural model for sentiment analysis. The performance of each segmentation method is evaluated under Convolutional Neural Network (CNN) and Recurrent Neural Network (RNN) model proposed in the literature for sentiment classification.
\end{abstract}

\section{Introduction}
\label{chapter:intro}

Social media data created by a huge number of individual users is very dynamic and varies to a great extent both in terms of size and type \citet{2016_Giachanou_Survey_of_Twitter_Sentiment_Analysis}. Since such data tend to have opinionated content, it urged the researchers to create means to automate the understanding of user sentiment on a particular topic. Sentiment analysis being about understanding author feeling, it largely focuses on positive or negative sentiment directed at an actor or an event. 
Earlier techniques on sentiment extraction have focused on the lexicon and rule-based solutions; however, studies recently started to employ Neural Networks. Even though latest studies show promising results, \citep{DBLP:journals/corr/0001KYS17} they largely focus on English and lack in-depth investigation for other languages.

For sentiment analysis there are two major challenges with most languages. First is that they mostly lack  well-known and useful resources such as Word-net, and there are few types of research on them. The other is that some of the languages are morphologically richer than the most others, which leads to a different set of problems.
Lack of NLP resources such as WordNet, \citep{Miller:1995:WLD:219717.219748} SentiWordNet
\citep{Esuli06sentiwordnet:a, Baccianella10sentiwordnet3.0}, and SenticNet \citep{Cambria:2014:SCC:2892753.2892763} is a problem in case of lexicon or rule-based approaches being used. However, usage of neural network approaches based on embeddings and representation learning can eliminate this effect to a great extent. 
%
The rich morphology, such as in Turkish, introduces a large vocabulary problem, which will effectively diminish neural network's effectiveness in extracting proper vector representations for words and building the corresponding embedding later. This is partly due to large vocabulary having the majority of tokens with very small frequencies within the text, and partly due to different variations of words particular affixes being used in certain contexts.
In addition to the large vocabulary problem, the informal texts also contribute to the problem with the introduction of free style usage of language with slang, typographical errors, abbreviations, and the local usage of language in different areas. Several prominent approaches that can tackle these problems are presented in this work. Morphological analysis, sub-word segmentation are some of these approaches that can deal with large vocabulary problem and unstructured usage of the language. 

%

In morphologically rich languages, token derivation depends heavily on grammar rules and affixes. This has two major consequences that bring additional challenges for sentiment extraction task. Firstly, the language has a very large vocabulary that causes sparsity and a high level of dimensionality when a bag of words representations are used. A large vocabulary is a problem for the neural models as the size of word embedding matrix increases. This causes neural network learn slower and less efficiently. Secondly, an affix can determine the entire sentiment of a sentence, thereby increasing the importance of correctly identifying sub-word elements that incorporate such sentiments. Table~\ref{table-suffix-example} shows good examples of simple suffixes having a major sentimental effect.

\begin{table}
\caption{Turkish words w/ and w/o negation}
\label{table-suffix-example} 
\centering
\begin{tabular}{ll}
\toprule
\textbf{Turkish}		 & \textbf{Translation}\\
\midrule
Seviyorum	 & I like \\ 
Sev\textbf{mi}yorum	 & I don't like \\ 
Konuşmaktayız	 & We are (in the act of) talking \\ 
Konuş\textbf{ma}yız	 & We don't talk \\ 
\bottomrule 
\end{tabular}
\end{table}

As already mentioned before, most languages mostly lack advanced resources such as Word-net and studies that mostly depend on frameworks that are still in the making or already abandoned. One such development is Zemberek, \citep{Akin2007} which is originally developed by \citeauthor{Akin2007} with further improvements until 2010\footnote{\url{github.com/ahmetaa/zemberek}}. The project was discontinued or seen a low level of development traffic for a while; however, it was still being widely used due to lack of resources to utilize otherwise. The project was re-adopted with a new code base\footnote{\url{github.com/ahmetaa/zemberek-nlp}} starting in late 2013 with further improvements to the date. 

In this work, we primarily focus on Turkish. This is because Turkish suffers both from large vocabulary due to its morphology and from lack of resources such as Word-net. Sentiment analysis for Turkish has been addressed using lexicon \citep{Erogul2009,Vural2012,kokciyan2013bounce,yildirim2015impact} or rule-based \citep{Boynukalin2012,uccan2014otomatik,Coban_2015} methods. 

To the best of our knowledge, this is the first work that reports a detailed empirical evaluation of a neural network based approach to sentiment analysis for Turkish. \citet{onal2015SocialMediaNer,demir2014improving,2016_kuru_charner} are examples of successful use of word embeddings and neural nets for Named Entity Recognition (NER). \citet{2016_kuru_charner} propose a character-level LSTM for the NER task. 
The main contribution of this work is investigating the effect of various segmentation models, including word-based, character-based, as well as morphological analysis based segmentation for sentiment classification. For the sake of completeness to the investigation, several different neural network models are also incorporated into the text classification pipe-lining. Together with the challenges of working with a morphologically rich language, another challenge is the use of social media resources, which enforces no formal grammar or language controls over the accuracy of the input texts. We evaluate each segmentation method with a CNN and an RNN model, which, together are two states of the art neural network models for sentiment analysis \citep{DBLP:journals/corr/0001KYS17}. The contributions of the work can be summarized as follows:

\begin{itemize}
	\item This work evaluates the effects of different segmentation methods on performance in terms of accuracy and computational effectiveness.
	
	\item It will also enable to compare different deep neural networks.
	
	\item To the best of our knowledge, it will be the first sentiment analysis experiment which applies Neural Network on Turkish data sets.
	
\end{itemize}



The organization of the paper is as follows.

\begin{description}
	\item[Related Work:] In this section, we present the state of the art sentiment analysis studies and experiments both for Turkish and other languages.  
	\item \textbf{\proposedwork{}:} In this section, the text segmentation methods and deep learning models, used in this work and the methodology used are described. 
	\item[Experiments and Results:] In this section, we present our settings order to execute processes involved in proposed methodology, experiments and the results.
	\item[Conclusion and Future Work:] Finally, we  conclude with an overview of the work and potential future extensions to this work.
\end{description}

\section{Related Work}
\label{chapter:rw}
\subsection{Neural networks for Sentiment Analysis}

In recent years, neural network models that can encode the sentiment of a text into a distributed representation have been studied intensively. Besides, studies that rely on pre-trained word embeddings such as the Bag of Semantic Concepts, \citep{2014_Lebret_N_gram_Based_Low_Dimensional_Representation_for_Document_Classification}, recursive neural networks \citep{SC:2012_Socher_Semantic_Compositionality_Through_Recursive_Matrix_Vector_Spaces,SC:2013_Socher_Recursive_Deep_Models_for_Semantic_Compositionality_Over_a_Sentiment_Treebank,SC:2014_Irsoy_Deep_Recursive_Neural_Networks_for_Compositionality_in_Language,SC:2014_Irsoy_Opinion_Mining_with_Deep_Recurrent_Neural_Networks} and CNNs \citep{SC:2014_Kalchbrenner_Convolutional_Neural_Network_for_Modelling_Sentences} have been utilized as neural semantic compositionality models for sentiment. In \citet{DBLP:journals/corr/0001KYS17} CNN and RNN and their usages are compared for NLP related tasks. 

In  \citet{2014_Lebret_N_gram_Based_Low_Dimensional_Representation_for_Document_Classification}, bag of words approach is used by extracting n-gram representations from words. The main motivation for breaking down individual words into smaller sub-components is manifested as a way of dealing with large vocabulary size. The authors also use K-means clustering to further deduct vector size for word representation. 

The work in \citet{SC:2014_Irsoy_Deep_Recursive_Neural_Networks_for_Compositionality_in_Language} propose a solution using a positional directed acyclic graph with a Recursive Neural Network\citep{SocherEtAl2011:RNN}. In their method, weight for a sentiment is calculated by extracting a weight for each token by propagating the weights of nodes in the binary tree that store the word vectors. 
Another example of using Deep Neural networks for sentiment and opinion extraction is provided in \citet{SC:2014_Irsoy_Opinion_Mining_with_Deep_Recurrent_Neural_Networks}. The authors propose using an Elman-type Recurrent Neural Network \citep{COGS:COGS203} with improved features such as adding neural depth by stacking Elman hidden layers \citep{NIPS2013_5166} and utilizing bi-directionality with Bidirectional RNN architecture proposed by \citet{Schuster:1997:BRN:2198065.2205129}. 
A major problem with word-vectors is the limitation on vector space definitions. In \citep{2014_Lebret_N_gram_Based_Low_Dimensional_Representation_for_Document_Classification, DBLP:journals/corr/abs-1003-1141}, and \citet{SC:2012_Socher_Semantic_Compositionality_Through_Recursive_Matrix_Vector_Spaces}, a Recursive Matrix-Vector Model (MV-RNN) is proposed in order to achieve compositionality. The model enables to learn compositional vector representations for phrases and sentences that constitute various types and lengths. The model achieves this by assigning a vector and a vector matrix to every node in the parse tree. The MV-RNN starts with building multi-word vectors by building multiples of single word vector representations using vectors from constituting words.

In \citet{SC:2015_Li_When_Are_Tree_Structures_Necessary_for_Deep_Learning_of_Representations}, the authors compare RNN and recursive neural models on the tasks of sentiment classification of sentences and syntactic phrases, question answering, discourse parsing, and semantic relations. They report that RNN models have an equal or superior performance to recursive models except for the semantic relations between nominals task.
\citeauthor{SC:2014_Kalchbrenner_Convolutional_Neural_Network_for_Modelling_Sentences} propose Dynamic Convolutional Neural Network (DCNN) model  \citep{SC:2014_Kalchbrenner_Convolutional_Neural_Network_for_Modelling_Sentences} in order to model semantic compositionality of sentences. The DCNN model is compared against recursive neural networks on Stanford Sentiment Tree-bank. \citep{SC:2013_Socher_Recursive_Deep_Models_for_Semantic_Compositionality_Over_a_Sentiment_Treebank} The DCNN model is shown to outperform the recursive neural network model of \citet{SC:2013_Socher_Recursive_Deep_Models_for_Semantic_Compositionality_Over_a_Sentiment_Treebank} on binary and multi-class sentiment classification. It is important to note that, in their work, DCNN model is not compared with RNN models.
CNN models proposed by~\citet{Kim_2014} for sentiment analysis on sentences outperform DCNN on 2 out of 3 datasets. \emph{CNN-rand} is one of those CNN model variants that does not use word2vec, which has lower scores than other 3 word2vec-powered variants to some extent.

\subsection{Sentiment Analysis for Turkish}

The survey on Turkish sentiment analysis given in \citet{2016_Dehkharghani_sentiment} provides a thorough analysis of sentiment analysis, and propose a system of methods to analyze Turkish in this context. The authors propose an approach to process Turkish texts at different granularity levels. Levels are proposed as \emph{Word-level}, \emph{Phrase-level}, \emph{Sentence-level}, and \emph{Document-level}. This work also elaborates on several linguistic issues such as negation, intensification, conditional sentences, rhetorical questions, sarcastic phrases, and idiomatic uses. In addition, it also raises some of the other issues such as emoticons, conjunctions, domain-specific indicative keywords, and background knowledge. T
The solution provided largely depends on SentiTurkNet \citep{2015_Dehkharghani_SentiTurkNet} framework, which is also developed by the same team.

One of the first studies on sentiment analysis in Turkish goes back to a thesis work by \citeauthor{Erogul2009} given in \citeyear{Erogul2009} where he applied a set of sentiment analysis processes to extract features from \datasetbeyazperde{} dataset. In this work, the author investigates the sentiment analysis performance  under different sets of features based on frequencies, the root of words, part-of-speech, and n-grams. 
In~\citet{Vural2012}, by using SentiStrength\footnote{\url{http://sentistrength.wlv.ac.uk/}}, Vural et al. applied sentence-binary, sentence-max/min, and word-sum  features for sentiment analysis on the same dataset. SentiStrength is a lexicon-based sentiment analysis library developed by~\citet{ASI:ASI21416}. 
In \citet{uccan2014otomatik}, sentiment analysis  performance on \datasetbeyazperde{} is further improved by including a feature selection step. 
 
In \citet{Demirtas_2013}, Demirtas and Pechenizkiy proposed a model for sentiment analysis by machine-translating Turkish texts, and evaluated the model on \datasetproducts{} data set. 
They both collect movie reviews from BeyazPerde\footnote{\url{http://www.beyazperde.com/}} having the same size with benchmark English movie review dataset. They also collect smaller datasets from multidomain product reviews from a Turkish online retailer\footnote{\url{http://www.hepsiburada.com/}} website. 
The study aimed to translate datasets into English and then classify texts using different ML algorithms including NB, SVM and Maximum ENtropy classifiers, after translation. 


\section{\proposedwork}
\label{chapter:pw}
In this work, it is aimed to investigate the effect of various segmentation methods on sentiment analysis by using well-known deep neural models on Turkish texts. Historically, Recurrent Neural Networks (RNNs) have been seen as the primary candidate for text classification \citep{SC:2012_Socher_Semantic_Compositionality_Through_Recursive_Matrix_Vector_Spaces,SC:2013_Socher_Recursive_Deep_Models_for_Semantic_Compositionality_Over_a_Sentiment_Treebank,DBLP:journals/corr/abs-1003-1141}. However, further recent solutions with Convolutional Neural Networks (CNNs) also provided at least as good results as RNNs. Both RNNs and CNNs have shown promising results \citep{DBLP:journals/corr/0001KYS17} for English sentences. 
We use these two neural models for sentiment analysis. As the segmentations methods, we experimented with \basemodelscount{} major methods that yield vocabularies with different size and characteristics. We present the details of the segmentation methods, the CNN-rand \citep{Kim_2014} and the LSTM neural models in what follows.

\subsection{Segmentation Methods}
\label{segmentation}
Segmentation helps divide each review into tokens. We primarily focused on~\basemodelscount{} major approaches for segmenting raw text into tokens, namely, \basemodels{} methods.
Among them, \emph{Word based} segmentation yields largest vocabulary, whereas the vocabulary is the set of characters with the \textbf{Character based} segmentation method. Besides the vocabulary size, the length of input sequences increases as the token granularity is shifted from words to characters. In principle, \baseword{} model yields the largest vocabulary with shortest sequences. The trend is towards smaller vocabulary and longer sequences while moving from this point to \basehybrid, \basemorph{} and finally \basesubword{}, respectively. 

We will illustrate each segmentation method with following sentences as the running example:

\begin{description}
	\item [Sentence 1:] film bastan sona duygu somurusu ama anlayan nerde!
	\begin{description}
		\item[English:] the movie \emph{exploits} (has typo) emotion from beginning to end, but who would care!
	\end{description}
	\item [Sentence 2:] geçen hafta elimize ulaştı, kullanımı kolay bulaşıkları pırıl pırıl yıkıyor.
	\begin{description}
		\item[English:] we got it delivered last week, it's easy to use and it washes dishes very well.
	\end{description}
\end{description}

\subsubsection{\baseword{} Segmentation}
The \baseword{} approach is the mostly used model in recent studies. This is both due to that it is easy to use and it provides reasonable accuracy results for non-additive languages such as English. Tokens in non-additive languages are mostly the same as word roots. However, this does not hold in case of additive languages such as Turkish. A root word could be converted to hundreds of distinct tokens just by using different suffixes,  which might extend vocabulary size to large numbers. In this work, we only deal with \modeltoken{} model in our experiments. 
For \baseword{} segmentation, \modeltoken{} can be obtained by simply using the tokenized text directly. Table~\ref{sample-segmentation-running-examples-word} shows \wordbasedlist{} variants' segmentations for the running example. 

\begin{table}[t!]
	\begin{threeparttable}
		\caption{Result for \baseword{} segmentation method on the running example}
		\label{sample-segmentation-running-examples-word}
		\begin{tabular}{ll}
			\toprule
			  \textbf{Segmentation}		 & \textbf{Output}\\
			\midrule
			\midrule
			 \modeltoken{} &  \pbox{10cm}{film bastan sona duygu somurusu ama anlayan nerde !} \\
			 \midrule
			 \modeltoken{} &  \pbox{10cm}{geçen hafta elimize ulaştı , kullanımı kolay bulaşıkları pırıl pırıl yıkıyor .} \\
			\bottomrule
		\end{tabular}
	\end{threeparttable}
\end{table}

\subsubsection{\basemorph{} Segmentation}

The \basemorph{} segmentation fragments each token into its building blocks by means of grammar rules for the language of the text. One challenge with the \basemorph{} method is developing, finding and using morphologic tools in an efficient and correct manner. Another challenge is to decide which information to keep and which to ignore. In this work, we used Zemberek \citep{Akin2007} for all morphological analysis related segmentations. We extracted embedded morphological information using several different approaches. 

\begin{table}[t!]
	\begin{threeparttable}
		\caption{The result for \basemorph{} segmentation method on the running example}
		\label{sample-segmentation-running-examples-morph}
		\begin{tabular}{ll}
			\toprule
			    \textbf{Segmentation}		 & \textbf{Output}\\
			\midrule
			\midrule
			 \modellemma{} &  \pbox{10cm}{film bas So duygu somurusu âmâ anlamak Ner !} \\
			 \midrule
			 \modellemma{} &  \pbox{10cm}{geçen hafta el ulaşmak , kullanım kolay bulaşık pırıl pırıl yıkamak .} \\
			 \midrule
			 \modellemmameta{} &  \pbox{10cm}{Noun A3sg Pnon Nom Noun A3sg Pnon Abl Noun A3sg P2sg Dat Noun A3sg Pnon Nom Unk Adj Adj PresPart Noun A3sg Pnon Loc Punc} \\
			 \midrule
			 \modellemmameta{} &  \pbox{10cm}{Adj Adv Noun A3sg P1pl Dat Verb Pos Past A3sg Punc Noun A3sg Pnon Acc Adj Noun A3pl P3pl Nom Dup Dup Verb Pos Prog A3sg Punc} \\
			 \midrule
			 \modellemmasuffix{} &  \pbox{10cm}{film A3sg Pnon Nom bas A3sg Pnon >dAn So A3sg In +yA duygu A3sg Pnon Nom somurusu âmâ anlamak +yAn Ner A3sg Pnon >dA !} \\
			 \midrule
			 \modellemmasuffix{} &  \pbox{10cm}{geçen hafta el A3sg ImIz +yA ulaşmak Pos >dI A3sg , kullanım A3sg Pnon +yI kolay bulaşık lAr I Nom pırıl pırıl yıkamak Pos Iyor A3sg .} \\
			 \midrule
			 \modelstem{} &  \pbox{10cm}{film bas so duygu somurusu ama anla ner !} \\
			 \midrule
			 \modelstem{} &  \pbox{10cm}{geçen hafta el ulaş , kullanım kolay bulaşık pırıl pırıl yık .} \\
			 \midrule
			 \modelstemmeta{} &  \pbox{10cm}{Noun A3sg Pnon Nom Noun A3sg Pnon Abl Noun A3sg P2sg Dat Noun A3sg Pnon Nom Unk Adj Adj PresPart Noun A3sg Pnon Loc Punc} \\
			 \midrule
			 \modelstemmeta{} &  \pbox{10cm}{Adj Adv Noun A3sg P1pl Dat Verb Pos Past A3sg Punc Noun A3sg Pnon Acc Adj Noun A3pl P3pl Nom Dup Dup Verb Pos Prog A3sg Punc} \\
			 \midrule
			 \modelstemsuffix{} &  \pbox{10cm}{film A3sg Pnon Nom bas A3sg Pnon >dAn so A3sg In +yA duygu A3sg Pnon Nom somurusu ama anla +yAn ner A3sg Pnon >dA !} \\
			 \midrule
			 \modelstemsuffix{} &  \pbox{10cm}{geçen hafta el A3sg ImIz +yA ulaş Pos >dI A3sg , kullanım A3sg Pnon +yI kolay bulaşık lAr I Nom pırıl pırıl yık Pos Iyor A3sg .} \\
			 \midrule
			 \modeltokenmeta{} &  \pbox{10cm}{Noun Noun Noun Noun Unk Adj Adj Noun Punc} \\
			 \midrule
			 \modeltokenmeta{} &  \pbox{10cm}{Adj Adv Noun Verb Punc Noun Adj Noun Dup Dup Verb Punc} \\
			\bottomrule
		\end{tabular}
	\end{threeparttable}
\end{table}

The full list of \basemorph{} segmentation variants is listed as follows: 
\begin{itemize}
\item In \modellemma{} approach, we extract lemma for each word. Rest of the word and suffixes are discarded.
\item In \modellemmasuffix{} approach, we also extract suffixes and concatenate them onto lemma as separate tokens. For suffixes lacking lexicon representation, the suffix class itself is used.
\item In \modellemmameta{} approach, we extract lemma positional attribute. In addition to this, we also extract suffix classes for all suffixes. We included this variant especially in order to observe whether word and suffix positional states hold significant information for sentimental content. 
\item In \modelstem{} approach, we extract stem for each word. Rest of the word and suffixes are discarded.
\item In \modelstemsuffix{} approach, we also extract suffixes and concatenate them onto stem as separate tokens. For suffixes lacking lexicon representation, the suffix class itself is used.
\item In \modelstemmeta{} approach, we extract lemma positional attribute. Additionally, we extract suffix classes for all suffixes. As in modellemmameta{} approach, we included this variant in order to  analyze whether word and suffix positional states hold significant information for sentimental content. 
\end{itemize}
It is important to note that in \modellemma{} and \modelstem{} approaches, the negation is also discarded since the negation is a suffix. However, if the word root itself intrinsically holds negation, the meaning is not effected.

For \basemorph{} segmentations, we first construct a dictionary containing all tokenized words used in datasets. Later on, a consumer processes the dictionary file to extract translation of each token into different representations for each of \morphlist{} models. Having this dictionary with the representation of each word integrated for each \basemorph{} model, we can now traverse sentences in each review and encode tokens inside each sentence into corresponding token sets. Table~\ref{sample-segmentation-running-examples-morph} shows \morphlist{} outputs for running examples. 

\subsubsection{\basesubword{} Segmentation}

The \basesubword{} segmentation breaks down each word into its building blocks without considering the underlying morphology. By breaking down words, the vocabulary size is reduced by a considerable amount depending on the approach or the parameters used. In this work, we elaborated on  three major sub-approaches under \basesubword{} segmentation. The first one is \emph{Byte-pair encoding}, which uses frequently used sub-word occurrences. The second one is \emph{n-gram}, which we use 1-gram variant. Finally, the last one is \emph{syllable based} segmentation. 

\emph{Byte-pair Encoding} (BPE) has been shown to be an effective way of dealing with large vocabularies in neural machine translation. The study in \citet{DBLP:journals/corr/SennrichHB15} proposes to use sub-word units such as morphemes and phonemes for neural machine translation. These sub-word units are extracted using BPE due to its robustness in automatically determining the morphemes and phonemes using the language corpora. The BPE method first builds a vocabulary from a corpus iteratively by merging the frequently co-occurring token pairs. The number of iterations is defined in advance and determines the size of vocabulary. In a second step, segmentation is performed by splitting the words into tokens using the vocabulary built in the first step. The parts of words that can be reconstructed by the vocabulary tokens are retained as individual tokens. The infrequent substrings in the text are broken down until they match a known token. If not possible, they are discarded. This ensures that the vocabulary size of output text remains within desired limits. Note that, in BPE, there is no limitation on the length of vocabulary tokens and the tokens are not required to be meaningful on their own. Since BPE requires a limit on vocabulary size for segmenting input text, we applied it with different vocabulary sizes. 1000, 5000 and 3000 vocabulary sizes are used for \bpelist{} approaches, respectively 
\footnote{For \bpelist{} models, we used the encoder given in \url{github.com/soaxelbrooke/python-bpe}}. 

In \emph{N-gram Character Segmentation}, text is broken down into sub-word elements with a maximum size of $n$ characters. For this work, we used 1-gram (a.k.a.,  \emph{character-based} segmentation, which simply breaks down words into standalone characters) as also applied in \citet{DBLP:journals/corr/LeeCH16}. 
The motivation behind this model basically comes from studies focusing on far-eastern languages such as Mandarin and Japanese. The alphabet for the majority of far eastern languages is made of self-identifying characters where each character has a distinct meaning. 

The third sub-word text segmentation method analyzed in this work is \emph{syllabification} or \emph{hyphenation} of text. Syllables can be automatically inferred from words in Turkish \citep{asliyan2007}. They are a core part of Turkish language and they are widely used \citep{L14-1375, asliyan2007, ccoltekin2007syllables} in Turkish NLP tasks. They also follow distinct patterns with a set of different forms due to deterministic nature of Turkish pronunciation, where each character almost exclusively represents the same sound or phoneme. Even though there are several different implementations of syllabification of Turkish texts, we preferred to implement our own solution by using the regular syllable forms are provided in \cite{asliyan2007}  \footnote{For \modelsyllable{} segmentation method, the code implemented within the scope of this work is available at  \url{github.com/ftkurt/python-syllable}} .  
Regular and irregular syllable forms along with character sets are as follows:

\begin{description}
	\item[Vowels (V):] a, e, ı, i, o, ö, u, ü
	\item[Consonants (C):] b, c, ç, d ,f, g ,ğ, h, j, k, l, m, n, p, r, s ,ş ,t ,v, y ,z 
	\item[Regular Syllable Forms:] V, VC, CV, CVC, VCC, CCV, CVCC, CCVC
	\item[Regular Syllable Examples:] e, ev, ve, ver, erk, bre, mart
	\item[Irregular Syllable Forms:] C{C+}V, C{C+}VC, VC{C+}, CVC{C+}, C{C+}VC{C+}
	\item[Irregular Syllable Examples:] brre, trren, üfff, oturr, krrakkk (typos, foreign words or representation of sounds)
\end{description}

\begin{table}[t!]
	\begin{threeparttable}
		\caption{The result for \basesubword{} segmentation methods on the running example}
		\label{sample-segmentation-running-examples-subword}
		\begin{tabular}{ll}
\toprule
\textbf{Segmentation}		 & \textbf{Output}\\
\midrule
\midrule
 \modelbpmini{} &  \pbox{10cm}{film ba st an sona du y g u s o m ur u su ama anla ya n nerde !} \\
 \midrule
 \modelbpmini{} &  \pbox{10cm}{geçen hafta el im iz e ul aş tı , ku ll anı mı kolay b ul aş ık ları p ır ıl p ır ıl yı kı yor .} \\
 \midrule
 \modelbpmid{} &  \pbox{10cm}{film bastan sona duygu so mu ru su ama anlayan nerde !} \\
 \midrule
 \modelbpmid{} &  \pbox{10cm}{geçen hafta eli miz e ulaştı , kullanımı kolay bul aşı kları pı rıl pı rıl yıkıyor .} \\
 \midrule
 \modelbplarge{} &  \pbox{10cm}{film bastan sona duygu so mur usu ama anlayan nerde !} \\
 \midrule
 \modelbplarge{} &  \pbox{10cm}{geçen hafta elimize ulaştı , kullanımı kolay bulaşıkları pırıl pırıl yıkıyor .} \\
 \midrule
 \modelchar{} &  \pbox{10cm}{f i l m b a s t a n s o n a d u y g u s o m u r u s u a m a a n l a y a n n e r d e !} \\
 \midrule
 \modelchar{} &  \pbox{10cm}{g e ç e n h a f t a e l i m i z e u l a ş t ı , k u l l a n ı m ı k o l a y b u l a ş ı k l a r ı p ı r ı l p ı r ı l y ı k ı y o r .} \\
  \midrule
  \modelsyllable{} &  \pbox{10cm}{film bas tan so na duy gu so mu ru su a ma an la yan ner de !} \\
   \midrule
 \modelsyllable{} &  \pbox{10cm}{ge çen haf ta e li mi ze u laş tı , kul la nı mı ko lay bu la şık la rı pı rıl pı rıl yı kı yor .} \\
\bottomrule
		\end{tabular}
	\end{threeparttable}
\end{table}

Table~\ref{sample-segmentation-running-examples-subword} shows segmentations generated by using  \subwordlist{} method on the running example. 

\subsubsection{\basehybrid{} Segmentation}

This segmentation method is a hybrid solution in the sense that \emph{word-based} and \emph{character-based} segmentation methods are used together resulting with a \textbf{\emph{Word-Character}~\modelhybrid{}} model. The words which are not recognized by Zemberek \citep{Akin2007} (i. e., unknown words) are being broken down to its characters, while the rest is kept as whole words. This approach is adapted from the \emph{dual decomposition} model given in  \citep{Wang_2014}, which models segmentation as an optimization problem for selecting whole-words or underlying characters as base tokens.

\begin{table}[t!]
	\begin{threeparttable}
		\caption{The result for \basehybrid{} segmentation method on the running example}
		\label{sample-segmentation-running-examples-hybrid}
		\begin{tabular}{ll}
			\toprule
			\textbf{Segmentation}		 & \textbf{Output}\\
			\midrule
			\midrule
			\modelhybrid{} &  \pbox{10cm}{film bastan sona duygu s o m u r u s u ama anlayan nerde !} \\
			\midrule
			\modelhybrid{} &  \pbox{10cm}{geçen hafta elimize ulaştı , kullanımı kolay bulaşıkları pırıl pırıl yıkıyor .} \\
			\bottomrule
		\end{tabular}
	\end{threeparttable}
\end{table}

For \basehybrid{} segmentation, the dictionary created for \basemorph{} analysis is reused. Representation provided by~\modeltokenmeta{} holds information about the type of the tokens. When a token is not recognized by our morphological analyzer,~\emph{Zemberek},~\modeltokenmeta{} will have the value of $Unk$. Since for \hybridlist{} model, dual decomposition requires using \modelchar{} based segmentation and \modeltoken{} for known words, \modelhybrid{} is simply extracted by compiling these two by also checking the value of \modeltokenmeta{} for each token. Table~\ref{sample-segmentation-running-examples-hybrid} shows the \hybridlist{} segmentation on the running example. 

\subsection{Neural Models for Sentiment Analysis}

CNNs and RNNs are reported to provide successful accuracy results for text classification tasks in the literature. Therefore we aimed to utilize both of them in our work. For CNN we use a special variant ,CNN-rand proposed by \citet{Kim_2014}. For RNN, we use a special form of RNN, which is called Long-Short Term Memory (LSTM) network. 

\subsubsection{Convolutional Neural Networks}
\label{chapter:pw:cnn-rand}

CNNs are \emph{Deep Learning} models designed to minimize the need for data pre-processing. They use a set of different multilayer perceptrons to achieve various types of non-linearity for modeling input layers to the next stages. CNNs are also known to be shift invariant, which means they tend to be behaving the same wherever the focal point for the occurrence of a feature of interest is.

\begin{figure}[t!]
	\centering
	\includegraphics[width=13cm]{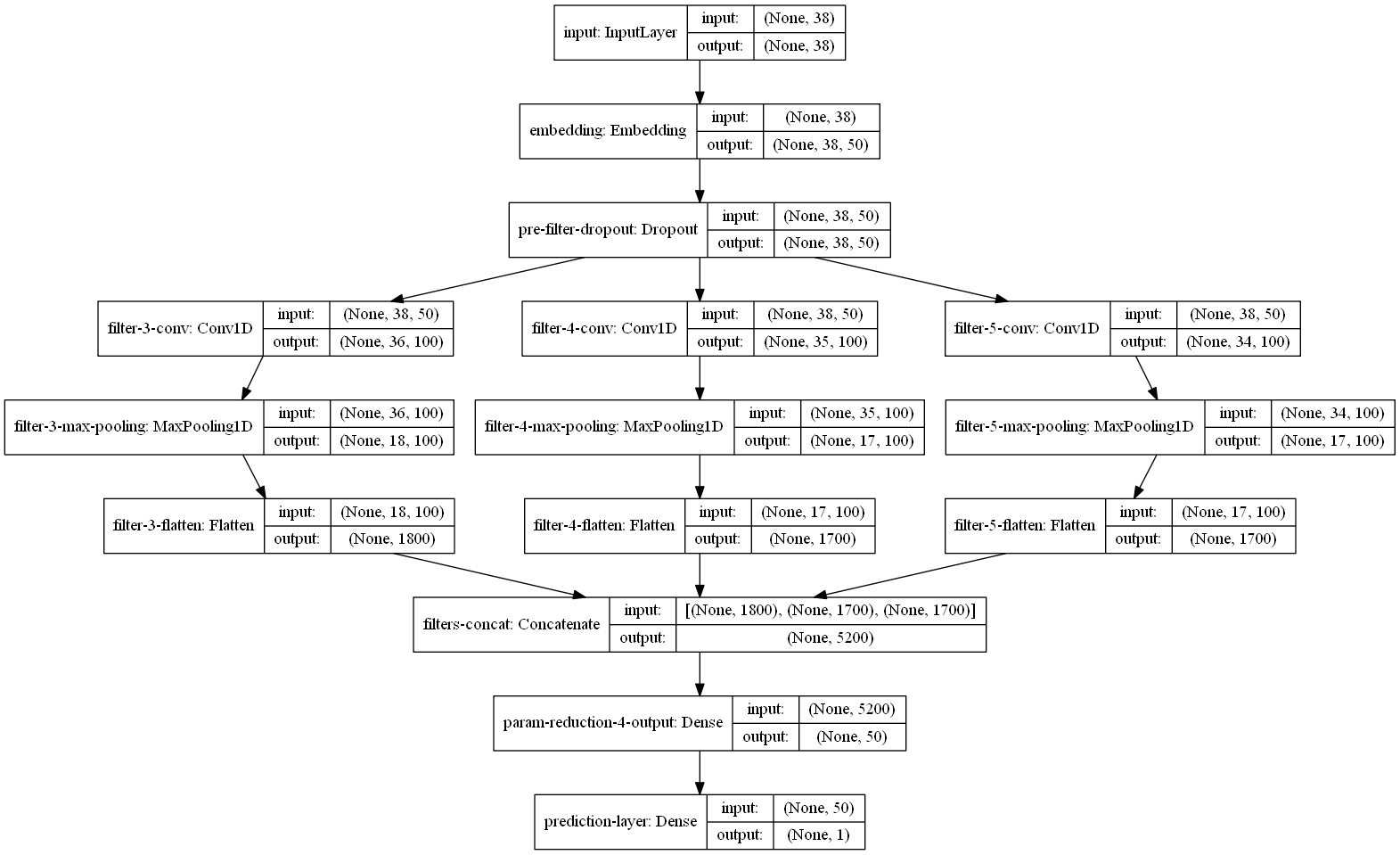}
	\caption{CNN-rand network described in \citet{Kim_2014} with layer shapes produced for \modeltoken{} segmentation on Movie Reviews dataset}
	\label{cnn-rand-model}
\end{figure}

We used one of CNN model variations described in \citet{Kim_2014} as our CNN model (See Figure~\ref{cnn-rand-model}). This variation initiates embedding with random values, instead of an external embedding such as \emph{word2vec \footnote{ Efficient Estimamtion of Word Representations in Vector Space, by T. Mikolov et al. \url{https://arxiv.org/abs/1301.3781}}}. We opted this model variation to minimize our dependence on a language corpus. Another reason is that  most of  our segmentation methods generate tokens that are not words.  
For this CNN model, we extended the source code of a publicly available implementation \footnote{\url{github.com/dennybritz/cnn-text-classification-tf}}. 
In this approach, data is represented through embeddings of tokens, in order to fit the resulting output to a vectorial space of equal sized batches, each review is then padded with special \textless PAD\textgreater{} tokens until it matches the longest entry.
Key layers introduced in this model (Figure~\ref{cnn-rand-model}) are described below.

\begin{description}
	\item[Input Layer:] This layer receives the input data both during training and evaluation. This layer has an input shape such as (None, 38) which indicates that every input sequence introduced has to have a length of 38, which is the maximum length for movie reviews dataset when processed with \modeltoken{} segmentation. The $None$ part indicates the batch size flexibility, which can be any number. 
	\item[Embedding Layer:] This layer keeps the vector representation of each token, in this case, words. The embedding size is introduced as 50, therefore the input shape attached by a depth of 50 at the output of the layer, which represents the vector size for tokens.
	\item[Dropout Layer:] Dropout layer is a layer which helps network learn seemingly insignificant patterns within data by randomly dropping a proportion of connections to the next layer. A dropout value of 0.5 is used for this layer. Therefore, it means that every time a data is introduced 50\% of connections from upper Embedding layer will be discarded and will not be passed onto lower layers. 
	\item[Filtering Layers:] This layer groups introduces a system filtering mechanisms to look deeper into phrases and groups of words inside input sequence. Paths are provided for filter sizes of 3, 4, and 5. Each path represents a filtering mechanism and has following sub-layers.
	\begin{description}
		\item[Convolution Layer:] This is the main convolution layer which introduces CNN part of the neural network. Convolutions are basically a smaller sized patch convolving over an input layer and passing the measured weights onto next layers. Next layers usually have smaller width and length and larger depth. Therefore, the convolutional layer translates spatial information into depth, which could be considered as a temporal output. For each of convolution layers in these filters, the patch sizes are selected as the size of filters. Therefore, the convolutions will be seeking for filter-sized frames (phrases). This setting helps to extract sentiment that is held by a group of words as opposed to the singular entries. Since the stride size is the filter size, the output layer is smaller by a factor of $stride-1$, i.e. an input size of 38 is converted to the output size of 36 for filter size 3. 
		\item[Max Pooling Layer:] This layer translates the input layer into smaller representations using Max pooling. In this case, the output is half the size of the input layer, which means each output parameter is obtained from the maximum value of corresponding 2 input parameters.
		\item[Flatten Layer:] This layer converts multi-dimensional parameter setting into single dimension. This is needed for concatenation layer at the end of filtering stages. In case of filter size of 3, an input layer of (None, 18, 100) is converted to (None, 1800), effectively reshaping input while keeping all parameters.
	\end{description}
	\item[Concatenation Layer:] This layer concatenates inputs from different layers and output them in a single shape. The output shape is simply a shape with a summation of input layers with no parameter being lost. In this case, input shapes of 1800, 1700, and 1700 are converted into an output shape of 5200.
	\item[Dense Layer:] This layer gradually reduces the number of parameters in order to achieve the prediction size. Sigmoid activation is used for this dense layer. In this case, the layer converts an input size of 5200 into an output of size 50. 
	\item[Dense Layer for Prediction:] This is the second dense layer which translates an input size of 50 into a single parameter which is needed for a binary sentiment analysis task as in our case. 
\end{description}

Key parameters for CNN models indicated by \citet{Kim_2014} are \emph{word embedding dimension}, \emph{dropout}, \emph{filter size} and a \emph{number of filters}.
The \emph{word embedding} represents the vectorial space for each vocabulary within input text. Dropout is a quite radical idea implemented in deep neural networks to let network better understand building blocks of the meaning the network is trying to extract from the input. \emph{Dropout} is the ratio of connections between two layers being randomly dropped during training. This lets network learn lower weight connections explaining input. \emph{Filter sizes} indicate the number of tokens to convolve over during training, and filters variable defines a number of filters for each filter size. 

\subsubsection{CNN-rand Simplified}
\label{chapter:pw:cnn-rand-simplified}

\begin{figure}[t!]
	\centering
	\includegraphics[width=14cm]{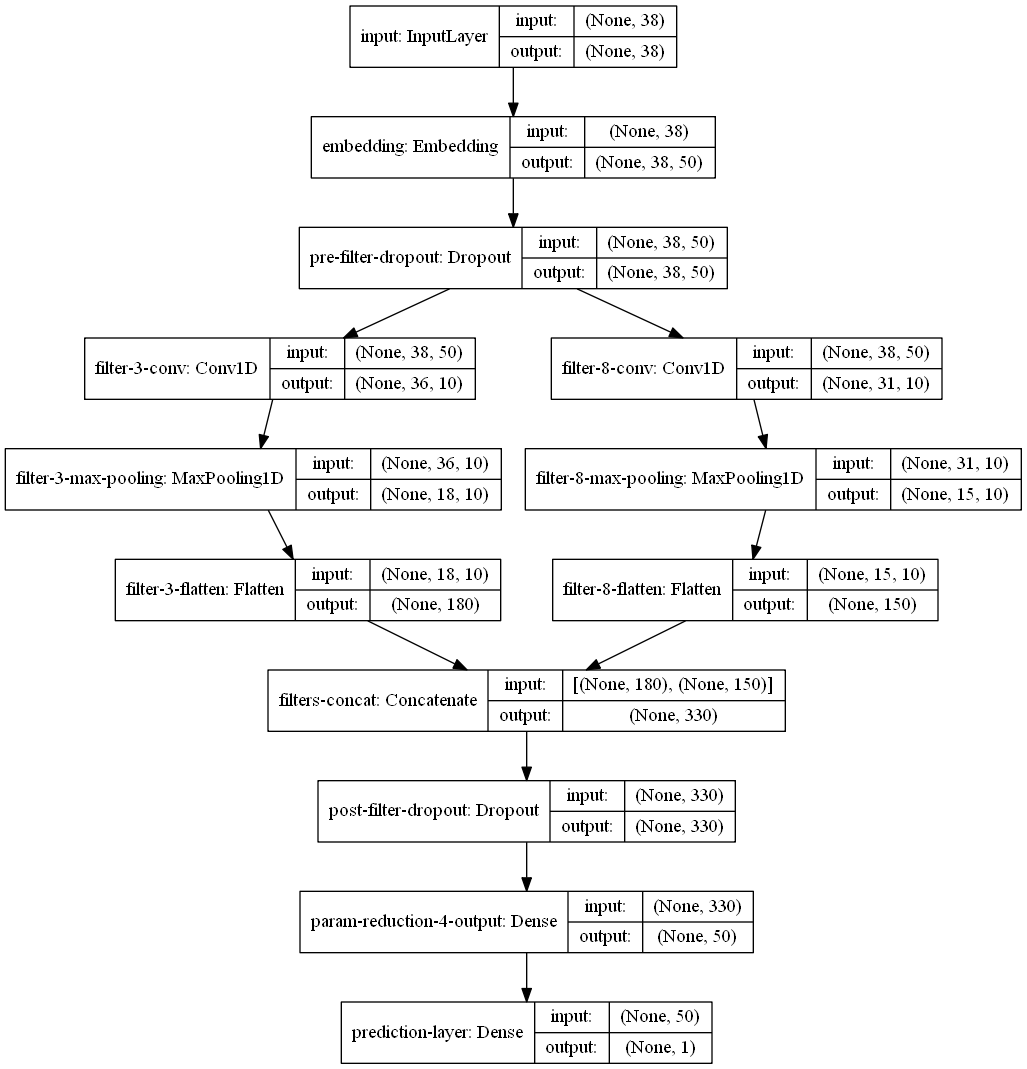}
	\caption{Simplified CNN-rand network with layer shaped for \modeltoken{} segmentation on Movie Reviews dataset}
	\label{simplified-cnn-rand-model}
\end{figure}

Simplified CNN-rand (Figure~\ref{simplified-cnn-rand-model}) is a smaller network version of the original CNN-rand network. It eliminates some of the filters and reduces sizes to a great extent. 
The changes introduced in this simplified version \footnote{\url{github.com/alexander-rakhlin/CNN-for-Sentence-Classification-in-Keras}} are as follows:

\begin{itemize}
	\item \emph{Smaller embedding dimension is used}: 20 instead of 300
	\item \emph{Fewer filter size is used}: 2 filter sizes instead of 3
	\item \emph{Fewer filters are used for each size. Proposes experiments showing that fewer is enough}: 3-10 filter size instead of 100
	\item \emph{Proposes random embedding initialization is no worse than word2vec init on IMDB corpus}: CNN-rand is preferred (as we already did)
	\item \emph{Network slides over Max Pooling instead of original Global Pooling.}
\end{itemize}

Key layer changes introduced in this model (Figure~\ref{simplified-cnn-rand-model}) are described below.

\begin{description}
	\item[Input Layer:] This layer is kept the same as the original.
	\item[Embedding Layer:] The embedding size for vector representations is reduced to 20 in this model.
	\item[Dropout Layer:] This layer is kept the same as the original. 
	\item[Filtering Layers:] Filter sizes are changed in this model. Filter sizes of 3, and 8 are used instead of the original sizes of 3, 4, and 5. Sub-layers for filters are used as in the original model.
	\item[Concatenation Layer:] This layer is kept the same as the original.
	\item[Dropout Layer:] A new dropout layer is introduced with a dropout value of 0.8 between concatenation layer and the dense later.
	\item[Dense Layer:] This layer is kept the same as the original.  
	\item[Dense Layer for Prediction:] This layer is kept the same as the original.  
\end{description}

\subsubsection{Long-Short Term Memory Neural Network}
\label{chapter:pw:lstm}
Long-Short Term Memory (LSTM) Neural Model is a special type of Recurrent Neural Network (RNN). In this RNN variant, a memory cell is incorporated into RNN cell that accumulates information throughout input propagation. This help network recognize input patterns over long intervals. LSTM was first proposed in \citet{Gers99learningto}, then the model was later investigated and improved in \citet{doi:10.1162/neco.1997.9.8.1735}. An LSTM cell consists of 4 main components: a cell, an input gate, an output gate and a forget gate. 

\begin{figure}[t!]
	\centering
	\includegraphics[width=6cm]{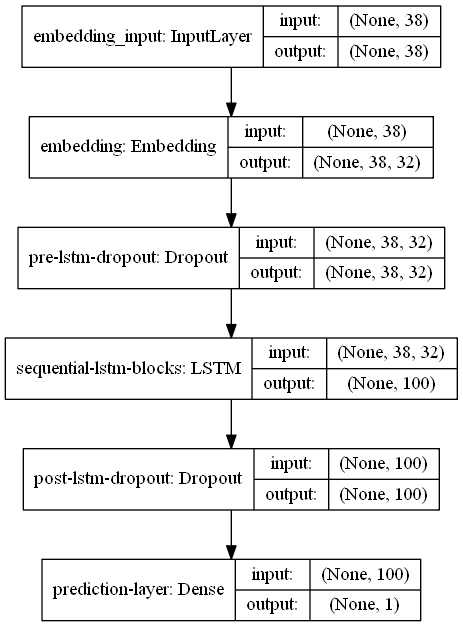}
	\caption{A simple LSTM network with layer shapes produced for \modeltoken{} segmentation on Movie Reviews dataset}
	\label{machine-learning-mastery-lstm-block}
\end{figure}

Long-Short Term Memory (LSTM) cells are designed for temporal data processing, and hence widely used in time series data. Each cell holds information about earlier tokens inside an entry and outputs a value based on current value and earlier tokens. In most cases, processing natural language is considered as a similar problem to the time series problem, since structure of the written or oral communication shows similarities to the time series data since both are sequential and temporal, as opposed to the image visual analysis.  Key layers introduced in the LSTM model used (Figure~\ref{machine-learning-mastery-lstm-block}) are as follows \footnote{ We extended our code given in \url{machinelearningmastery.com/sequence-classification-lstm-recurrent-neural-networks-python-keras/}}:

\begin{description}
	\item[Input Layer:] This layer receives the input data both during training and evaluation. The layer attributes are the same as attributes described for CNN models. 
	\item[Embedding Layer:] For this layer, the embedding size of 32 is introduced, therefore the input shape is attached by a depth of 32 at the output of the layer, which represents the vector size for tokens.
	\item[Dropout Layer:] A dropout value of 0.2 is used for this layer. Therefore, it means that every time a data is introduced 20\% of connections from upper Embedding layer will be discarded and will not be passed onto lower layers. 
	\item[LSTM Blocks:] This layer introduces LSTM blocks for sequence processing. It helps to extract sentiment using memory, processing, forgetting mechanisms built into the LSTM cells. 
	\item[Dropout Layer 2:] Another dropout layer is introduced after LSTM blocks for this model, and a dropout value of 0.2 is used.  
	\item[Dense Layer for Prediction:] This layer translates an input size of 100 into a single parameter which is needed for a binary sentiment analysis task as in our case. 
\end{description}

\subsubsection{Hyper-parameter Optimizations}

Deep learning hyper-parameters provide fine tuning the capabilities. In \citet{DBLP:journals/corr/ZhangW15b} it is indicated that they already used a grid search for hyper-parameter tuning and that selected parameters are identified as being the most effective. 
We performed a grid search for hyper-parameters to validate the effectiveness of the chosen parameters in \citet{DBLP:journals/corr/ZhangW15b} on the face of changing dataset and segmentation methods with the following value sets: 

\begin{description}
	\item[Filter Sizes:] (3,4,5), (10,16,22), (16,22,27), (22,27,33)
	\item[Dropout:] 0.4, 0.5, 0.6
	\item[L2 Regularization:] 0.0, 0.001, 0.01, 0.1 
\end{description}

Filter sizes are determined using number of characters statistics on our combined datasets. The number of tokens per each word is highest for \modelchar{} segmentation method. Therefore, we decided to use statistics with this segmentation method. The average number of characters in a word is $5.49$. Therefore, we used a linear space between original filter sizes (3, 4, 5) to around 5.5 times these values. However, after the first filter size set,  the difference between concurrent filter sizes is limited to ensure that the model does not lose any information for phrases with the length in between. 

We trained a CNN-rand model with \modelchar{} segmentation output for book reviews dataset with varying values of hyper-parameters provided earlier. Filter sizes used in testing are derived from the average number of characters in each word within the dataset. 

\begin{figure}[t!]
	\centering
	\includegraphics[width=13cm]{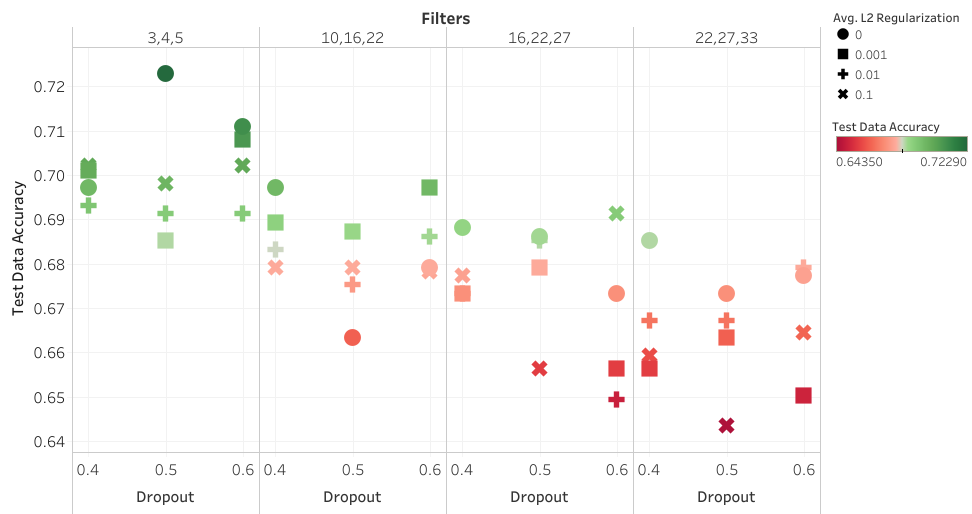}
	\caption{Accuracy graph for Dropout broken down by Filters.  Color shows Accuracy.  Shape shows details about L2 Regularization.}
	\label{hyper-parameter-tuning}
\end{figure}

Figure~\ref{hyper-parameter-tuning} indicates that 0.5 dropout value and 3,4,5 filter size set, which is the default parameter set for original work, derive the best results. It also shows that L2 adds no positive effect to the performance. As a result, for the experiments, we set dropout parameter to 0.5 and filter sizes to $3,4,5$. We did not use any L2 regularization and left it at the default value of 0. We used the default levels used recommended in the literature for other parameters.

\subsubsection{Model Training Callbacks}
\label{chapter:pw:callbacks}

For deep learning model building, we used Keras\footnote{\url{github.com/fchollet/keras}} with a Tensorflow\footnote{\url{github.com/tensorflow/tensorflow}} back-end. In order to build a neural network, the training data is presented to the model repetitively. This lets model better grasp training data and learn new things after model decides to do something different during an earlier repetition. This is a common practice and each iteration is called an \emph{Epoch}. At the end of each epoch, deep learning framework (Tensorflow) and the interface (Keras) provide access to run a set of functions before deciding if it is desired to continue with more epochs. These functions are called \emph{Callbacks}.
During our model building, we used following callbacks:

\begin{description}
	\item[BestModelSave:] This callback tracks some parameters and saves model when the best value so far is encountered. The variables to be tracked in order to save model is indicated when the callback is created.	
	\item[EarlyStopping:] When model training starts a constant epoch number should be provided. However, we cannot know for sure how many epochs will be sufficient for the model to learn training data. Therefore, as a rule of thumb, we provide a relatively high number of Epochs in order to make sure we trained the model sufficiently. However, if not interrupted, the model will most definitely over-fit to the data. Over-fitting is as bad as under-fitting; therefore, it is generally a good idea to stop training before that happens. Early stopping tracks variables of choice to stop training when configured monitors indicate that over-fitting started. \emph{EarlyStopping} needs monitoring parameter, minimum delta, and patience. Callback monitors the monitoring variable and constantly expects this parameter getting better. When it does not get better beyond minimum delta value it deducts from patience value. If the patience value reaches 0, training is interrupted. Every time a better result for monitoring variable is encountered, the patience value is reset. 
	\item[Progress Monitor:] These monitors mostly plot progress with training, epochs, accuracy, loss and expected remaining time. 
\end{description}

We used \emph{Validation Accuracy} as \emph{BestModelSave} monitoring parameter. Therefore, whenever the model achieves the highest accuracy on \emph{validation so far}, this callback saves the model.
For all models, we used validation loss as \emph{EarlyStopping} monitoring parameter, and 0.001 as minimum delta. Due to underlying differences between LSTM and CNN networks, we used 2, and 20 as patience value ,respectively. LSTM networks need fewer epochs (1-3) to fully train model. We used 200, and 5 as the number of Epochs for CNN, and LSTM networks, respectively.

\subsection{Limitations}
This work aims to demonstrate effective ways of dealing with sentiment extraction task for informal Turkish text with the least possible interference on input data. Therefore, this work does not utilize advanced Natural Language Processing (NLP) tools. Hence, operations listed below are not covered within this work. \begin{description}
	\item[Typo Checker:] This work does not perform typographical error checking and correction. The work aims to analyze the performance of different segmentation methods on informal texts. Utilizing such tools could have clouded our results from fully understanding the effects of studied segmentation methods and utilized neural networks.
	
	\item[Normalization:] Datasets include a fair amount of information encoded in different representations such as numbers, emoticons, shapes, images and etc. Even though handling these issues are shown to contribute to the results positively, we decided to use the raw form of input texts and propose this as future work.
	
	\item[Disambiguation:] Another shortcoming of morphological analysis demonstrated in this work is its inability to extract the correct form of each word and use the accurate roots and suffixes. However, this requires a deep understanding of language structures and grammar. It also requires the implementation of such advanced tools. This work does not rely on availability of such advanced tools, using them would increase the processing complexity.
	
	\item[Sentence Modelling:] It is also possible to model sentence structures and use this information to train more effective models for sentiment analysis. However, this will also add another layer of complexity to this work.
	
	\item[Global Word Vectors:] In this work, a set of different segmentation methods are utilized, and experiments for these methods are conducted on different datasets. Therefore, following shortcomings can be emphasized for this decision.
	
	\begin{itemize}
		\item Global word vectors defines vectors for whole words, but this work focuses on breaking down words into sub-word elements by various segmentation methods. The resulting tokens will not have any representation within these ready to use vector definitions. The available ones are very likely to be pointing to the incorrect vectors. For example, in a case where \emph{kamuflaj} (camouflage) is broken down to \emph{kamu} (public) + \emph{flaj} (no meaning) the resulting vector for \emph{kamu} will point out to the inaccurate vector representation. 
		
		\item Being a morphologically rich language creates scarcity for a very large number of rarely used word-suffix sets in Turkish. Hence, scarce tokens even if available within the WordVec library might have inaccurate representation due to lack of suitable sample text to train the vectors in a correct way.
		
	\end{itemize} 
	
\end{description}

\section{Experiments and Results}
\label{chapter:exp}
\subsection{Data Sets}
In the experiments, we used two benchmark datasets. \datasetbeyazperde{} is a collection of movie reviews from the Turkish movie platform Beyazperde\footnote{\url{www.beyazperde.com}}. The dataset contains 54K annotated paragraph-length reviews originally used in \citet{Erogul2009}. The dataset is collected from Turkish movie review site Beyazperde from various movies at random. Beyazperde users can rate movies on the scale of [0.5-5]. The dataset entries are labeled as follows. Reviews rated as 4.0-5.0 are accepted as positive, 2.5-3.5 as neutral and 0.5-2.0 as negative reviews.  The number of positive and negative instances for \datasetbeyazperde{} data set  is given Table~\ref{table-stats-beyazperde}.

\begin{table}[t!]
\centering
\caption{\datasetbeyazperde{} data set number of instances}
\label{table-stats-beyazperde}
\begin{tabular}{lcc}%
Dataset 			& Negative & Positive \\ \hline
Reviews 			& 27.000 & 27.000 \\ \hline
\end{tabular}
\end{table}

\begin{table}[t!]
\centering
\caption{\datasetproducts{} data set number of instances}
\label{table-stats-products}
\begin{tabular}{lcc} 
Dataset 			& Negative & Positive \\ \hline\hline
Movie 				& 5.331 & 5.331 \\ 
Book 				& 700 & 700 \\ 
Electronics 		& 700 & 700 \\
Kitchen 			& 700 & 700 \\
DVD  				& 700 & 700 \\ \hline
\end{tabular}
\end{table}

The \datasetproducts{} data set, which was used by the study in \citet{Demirtas_2013} was compiled from online retailer websites. It consists of 5 different sub datasets. In \citet{Demirtas_2013}, authors aim to improve text classification task for Turkish text by using machine translation. Therefore they collect reviews from Beyazperde by the same amount as their benchmark IMDB review dataset, which is in English. In addition to this movie reviews dataset, they also compile smaller datasets by collecting reviews for various products from online retailer Hepsiburada.com. The reviews are compiled for four different product categories, namely electronics, DVDs, kitchen products, and books in addition to the movie review dataset. For simplicity, we will be calling the full set of these 5 datasets as \datasetproducts. 
During dataset compilation, each review is labeled based on the rating user provided for the product upon review. Since the majority of the votes are $3+$ on a scale from 1 to 5, classification is exerted as $1-3$ rates being negative and $4-5$ being positive. The number of instances per label in \datasetproducts{} dataset is as given in Table~\ref{table-stats-products}.

\begin{table}[t!]
\centering
\begin{threeparttable}
\caption{Accuracy results on the \datasetproducts{} data set}
\label{table-results-products} 
\begin{tabular}{lccccc}
\toprule
\textbf{Model} & \textbf{Movie} & \textbf{Book} & \textbf{DVD} & \textbf{Electronics} & \textbf{Kitchen}\\
\midrule
Naive Bayes$^1$ 		 & 69.50 & \textbf{72.40} & \textbf{76.00} & \textbf{73.00} & \textbf{75.90}  \\ 
Naive Bayes MT$^1$ 		 & \textbf{70.00} 	 & 71.70 	 & 74.90 	 & 64.40 	 & 69.60  \\ 
Linear SVC$^1$ 			 & 66.00 	 & 66.60 	 & 70.30 	 & 72.40 	 & 70.00  \\ 
Linear SVC MT$^1$ 		 & 66.50 	 & 66.90 	 & 67.60 	 & 64.40 	 & 67.30  \\ 
\midrule
CNN @~\modellemmasuffix{} &  90.61 &  \textbf{81.43} &  76.43 &  77.86 &  75.00 \\
CNN @~\modelbpmid{}               &  90.61 &  75.71 &  76.43 &  75.00 &  \textbf{80.00} \\
CNN @~\modelhybrid{}                &  89.01 &  75.71 &  75.00 &  \textbf{80.71} &  77.14 \\
LSTM @~\modeltoken{}        &  \textbf{90.80} &  77.86 &  76.43 &  \textbf{80.71} &  75.00 \\
LSTM @~\modelbpmid{}               &  89.86 &  75.71 &  75.71 &  79.29 &  75.00 \\
LSTM @~\modellemma{}           &  89.20 &  80.00 &  \textbf{77.86} &  75.00 &  75.00 \\
\bottomrule 
\end{tabular}
\begin{tablenotes}
	\small
	\item $^1$\citet{Demirtas_2013}
\end{tablenotes}
\end{threeparttable}
\end{table}

\begin{figure}[t!]
	\centering
	\includegraphics[width=11cm]{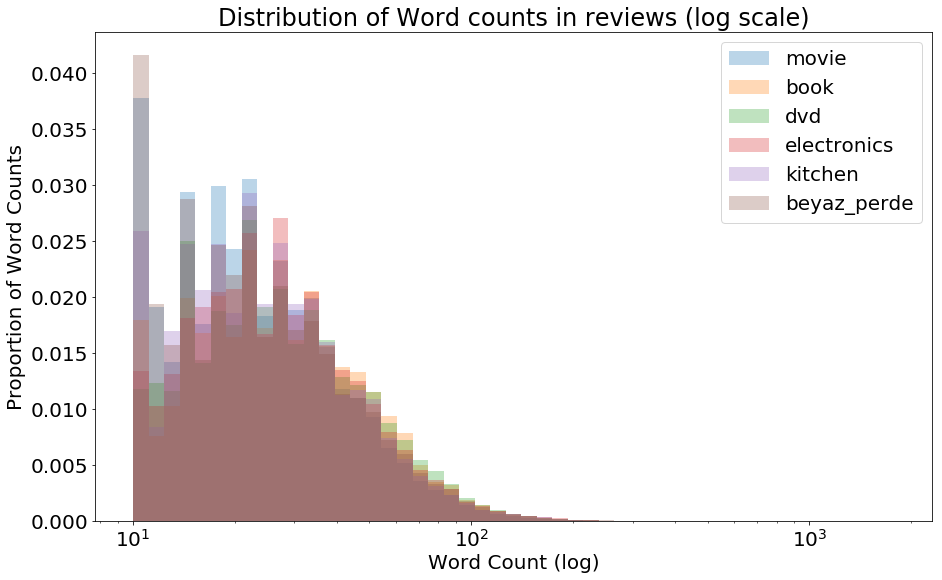}
	\caption{Histogram for the distribution of review length for datasets.}
	\label{histogram-datasets-review-length}
\end{figure}

In text classification task, particularly when working with machine learning methodologies, vocabulary size and data geometry are very important. Therefore, we also obtained basic statistics on datasets. Vocabulary size, average sentence length and maximum word count per review is provided in Table~\ref{table-dictionary} for both \datasetbeyazperde{} and \datasetproducts data sets. 
As another characteristics, the distribution of review length in terms of words seem to be varying to a great extent for various datasets. Figure~\ref{histogram-datasets-review-length} shows the differences in the distribution of sentences among different datasets.

\begin{table}[t!]
	\centering
	\caption{Vocabulary sizes and other basic statistics for \datasetbeyazperde{} and \datasetproducts{} datasets.}
	\label{table-dictionary}
	\begin{tabular}{lrrr} 
	{} &  Avg. Sentence Length &  Max Review Size &  Vocabulary \\
	\midrule
	beyaz\_perde.neg &         3.76 &    801 &    108,682 \\
	beyaz\_perde.pos &         3.73 &   1,717 &    112,566 \\ \hline
	book.neg        &         3.69 &    235 &      7,367 \\
	book.pos        &         3.51 &    126 &      5,634 \\
	dvd.neg         &         3.41 &    293 &      7,553 \\
	dvd.pos         &         3.30 &    289 &      6,564 \\
	electronics.neg &         3.91 &    281 &      7,791 \\
	electronics.pos &         3.58 &    245 &      5,904 \\
	kitchen.neg     &         3.58 &    130 &      6,475 \\
	kitchen.pos     &         3.44 &    210 &      5,457 \\
	movie.neg       &         2.36 &     59 &     19,481 \\
	movie.pos       &         2.41 &     59 &     17,825 \\\hline
\end{tabular}
\end{table}

\subsection{Data Pre-processing}
In the pre-processing phase, each review is firstly split into its sentence components using sentence detection functions provided by Zemberek, \citep{Akin2007} and sentences are tagged as their parent (the review the sentence is extracted from). In the second phase, each sentence is processed with different segmentation methods as discussed in Section \autoref{chapter:pw}. The main motivation behind this is the fact that neural networks are known to perform better with single sentence entries, due to the fact that a sentence mostly contains one distinct sentiment. Whereas, multiple sentences, even if they are in the same review, could contain different sentiments. Use of sentence, in this regard, is also a common practice for deep learning model training for sentiment analysis in English. In order to handle too long sentences, we decided to further forcefully break these entries down into smaller entries. To this aim, applied a filter with $percentile = 99.5$ to ensure that only sentences within longest $0.5\%$ of entries will be broken down. 

In segmentation stage, we created global corpus by compiling all dataset contents. This corpus is then used to create, build and use various encoders for different segmentation methods. These segmentation methods and involved processes are detailed in Section \autoref{segmentation}. Table~\ref{sample-segmentation-vocabulary-and-identification} shows vocabulary sizes for each segmentation method. However, it seems the vocabulary size is still exceptionally large for some of the segmentation methods. This is particularly true in case of \modeltoken{} which has a vocabulary size of $198k$. In order to overcome this large vocabulary problem and possibly reduce the vocabulary size further, we applied filtering by removing scarce words having frequency less than 3. During experiments, we used training/validation/test split with ratios of 0.8, 0.1, and 0.1, respectively. Table~\ref{table-train-val-test-split} shows the distribution of sentences into samples for each dataset. 

\begin{table}[t!]
	\centering
	\begin{threeparttable}
		\caption{Compiled vocabulary sizes for each segmentation methods}
		\label{sample-segmentation-vocabulary-and-identification} 
		\begin{tabular}{rc}
			\toprule
			\textbf{Segmentation}		 & \textbf{Vocabulary Size}\\
			\midrule
			\modelbpmini{}               &     957 \\
			\modelbplarge{}              &   28.676 \\
			\modelbpmid{}               &    4.822 \\
			\modelchar{}            &     112 \\
			\modelhybrid{}               &  117.294 \\
			\modellemma{}           &   99.872 \\
			\modellemmameta{}   &      96 \\
			\modellemmasuffix{} &   99.972 \\
			\modelstem{}              &   99.766 \\
			\modelstemmeta{}      &      96 \\
			\modelstemsuffix{}    &   99.863 \\
			\modeltoken{}        &  198.262 \\
			\modeltokenmeta{}        &      15 \\
			\bottomrule 
		\end{tabular}
	\end{threeparttable}
\end{table}

\begin{table}[t!]
	\centering
	\caption{For each data set, number of sentences in training, testing and validation partitions}
	\label{table-train-val-test-split}
	\begin{tabular}{lccc}
		Dataset 			& Train & Test & Validation \\ \hline\hline
		Beyaz Perde			& 162.561 & 20.393 & 20.467 \\ \hline
		Book 				& 4.001 & 515 & 548 \\
		Movie 				& 20.473 & 2.530 & 2.520 \\
		Electronics 		& 4.278 & 466 & 528 \\
		Kitchen 			& 3.961 & 493 & 485 \\
		DVD  				& 3.789 & 463 & 468 \\ \hline
	\end{tabular}
\end{table}

\subsection{Baseline Methods}

In this work, we used baseline scores for \datasetbeyazperde{} and \datasetproducts{} data sets from previous work. For \datasetbeyazperde{} data set we have 5 different scores from 2 different types of methods reported in \citet{Vural2012,uccan2014otomatik}. For \datasetproducts{}, we have 4 different scores reported in \citet{Demirtas_2013}. Baseline scores for \datasetbeyazperde{} dataset are listed in Table~\ref{beyaz-perde-baseline}. \citet{uccan2014otomatik} improves~\citet{Vural2012}'s accuracy results on \datasetbeyazperde{} by employing Support Vector Machine(SVM) and Naive Bayes Classifier(NB) and also by including a feature selection step. They employ two main methods for feature selection stage. Results they acquire by training an SVM and an NB classifier with features selected by both Chi-Square and Information Gain are also listed in our baseline scores for \datasetbeyazperde{} data set.

\begin{table}[t!]
\centering
\begin{threeparttable}
\caption{Baseline accuracy results on the \datasetbeyazperde{} data set}
\label{beyaz-perde-baseline} 
\begin{tabular}{llc}
\toprule
\textbf{Model}		 & Author & \textbf{Acc.}\\
\midrule
Sentence-binary	 & \citet{Vural2012}	& 70.39 \\ 
Sentence-max/min	 & \citet{Vural2012}	& 74.83 \\ 
Word-sum 			 & \citet{Vural2012}	& 75.90 \\ 
Chi-Square			 & \citet{uccan2014otomatik}	& \textbf{83.90} \\ 
Information Gain	 & \citet{uccan2014otomatik}	& \textbf{83.90} \\ 
\bottomrule 
\end{tabular}
\end{threeparttable}
\end{table}

\begin{table}[t!]
\centering
\begin{threeparttable}
\caption{Baseline accuracy results on the \datasetproducts{} data set}
\label{product-review-baseline} 
\begin{tabular}{lccccc}
\toprule
\textbf{Model} & \textbf{Movie} & \textbf{Book} & \textbf{DVD} & \textbf{Electronics} & \textbf{Kitchen}\\
\midrule
Naive Bayes$^1$ 		 & 69.50 & \textbf{72.40} & \textbf{76.00} & \textbf{73.00} & \textbf{75.90}  \\ 
Naive Bayes MT$^1$ 		 & \textbf{70.00} 	 & 71.70 	 & 74.90 	 & 64.40 	 & 69.60  \\ 
Linear SVC$^1$ 			 & 66.00 	 & 66.60 	 & 70.30 	 & 72.40 	 & 70.00  \\ 
Linear SVC MT$^1$ 		 & 66.50 	 & 66.90 	 & 67.60 	 & 64.40 	 & 67.30  \\ 
\bottomrule 
\end{tabular}
\begin{tablenotes}
	\small
	\item $^1$\citet{Demirtas_2013}
\end{tablenotes}
\end{threeparttable}
\end{table}

For \datasetproducts{} data set, results reported in \citet{Demirtas_2013} are used as baseline. Results for three different supervised learning algorithms, Naive Bayes, Support Vector Machine, and Maximum Entropy classifier, are reported, as given in Table~\ref{product-review-baseline}.

\subsection{Experiment Results and Discussion}

During experiments, we applied 13 different segmentation methods for 6 different datasets. We finally executed classification modeling on 3 different deep learning models,  CNN-rand, Simplified CNN-rand, and LSTM. In addition to sentence level sentiment analysis result, we obtained review level sentiment analysis results, as well. To this aim, in order to obtain review level sentiment, majority voting is applied for the sentences belonging to the same review. 

\subsubsection{CNN-rand Results}

\begin{table}[t!]
	\resizebox{\columnwidth}{!}{
	\centering
	\begin{threeparttable}
		\caption{CNN-rand accuracy results for sentence level sentiment analysis}
		\label{table-cnn-tf-results} 
\begin{tabular}{llllllll}
\toprule
{} &  Movie &   Book &    DVD & Electr. & Kitchen & Beyaz Perde & Avg \\
\midrule
\modeltoken{}        &  82.29 &  68.54 &  65.23 &  \textbf{74.03} &  68.36 &  \textbf{74.07} &  \textbf{72.09} \\
\modelhybrid{}                &  82.17 &  68.35 &  64.15 &  73.39 &  66.73 &  73.31 &  71.35 \\
\modellemma{}           &  81.78 &  67.77 &  62.42 &  72.75 &  \textbf{70.18} &  72.72 &  71.27 \\
\modellemmasuffix{} &  83.16 &  69.32 &  64.36 &  69.10 &  67.95 &  73.60 &  71.25 \\
\modelbpmid{}               &  82.06 &  66.41 &  64.79 &  73.39 &  66.73 &  72.57 &  70.99 \\
\modelsyllable{}             &  77.83 &  69.32 &  62.63 &  72.96 &  69.57 &  72.43 &  70.79 \\
\modelstem{}              &  82.29 &  \textbf{69.51} &  63.28 &  68.88 &  67.75 &  72.76 &  70.75 \\
\modelbplarge{}              &  \textbf{83.32} &  69.32 &  61.77 &  72.75 &  61.05 &  74.02 &  70.37 \\
\modelbpmini{}               &  78.46 &  65.05 &  \textbf{68.03} &  66.09 &  60.45 &  71.82 &  68.32 \\
\modelstemsuffix{}    &  82.65 &  68.74 &  64.15 &  46.78 &  68.97 &  73.18 &  67.41 \\
\modelchar{}            &  74.70 &  65.05 &  57.24 &  63.73 &  60.85 &  68.49 &  65.01 \\
\midrule
\modellemmameta{}   &  60.40 &  60.39 &  57.88 &  60.94 &  58.01 &  59.32 &  59.49 \\
\modelstemmeta{}      &  62.65 &  55.15 &  60.04 &  59.66 &  58.62 &  59.49 &  59.27 \\
\modeltokenmeta{}        &  58.62 &  57.28 &  55.51 &  54.94 &  52.33 &  54.31 &  55.50 \\
\bottomrule
\end{tabular}
\end{threeparttable}
}
\end{table}

We present our sentence level CNN-rand results in Table~\ref{table-cnn-tf-results}. Table shows that \modeltoken{} holds the highest average score of 72.09\%. It is followed by \modelhybrid{}, \modellemma{}, and \modellemmasuffix{} with almost equivalent accuracy values. We also present review level CNN-rand results in Table~\ref{table-cnn-tf-results-mv}. In this table, \modeltoken{} holds the highest average score of 80.81\%. It is followed by \modelbpmid{}, and \modelhybrid{} with similar scores. 

\begin{table}[t!]
	\resizebox{\columnwidth}{!}{
	\centering
	\begin{threeparttable}
		\caption{CNN-rand accuracy results for review level sentiment analysis}
		\label{table-cnn-tf-results-mv} 
\begin{tabular}{llllllll}
\toprule
{} &  Movie &   Book &    DVD & Electr. & Kitchen & Beyaz Perde & Avg \\
\midrule
\modeltoken{}        &  90.52 &  76.43 &  74.29 &  \textbf{79.29} &  76.43 &  \textbf{87.91} &  \textbf{80.81} \\
\modelbpmid{}               &  90.14 &  80.00 &  73.57 &  75.71 &  \textbf{78.57} &  86.11 &  80.68 \\
\modelhybrid{}                &  \textbf{90.80} &  77.14 &  72.86 &  77.86 &  76.43 &  86.91 &  80.33 \\
\modellemma{}           &  89.11 &  75.00 &  72.14 &  78.57 &  78.57 &  85.98 &  79.90 \\
\modellemmasuffix{} &  90.61 &  77.86 &  \textbf{75.71} &  71.43 &  72.14 &  87.19 &  79.16 \\
\modelsyllable{}             &  86.10 &  \textbf{80.71} &  68.57 &  75.71 &  77.14 &  86.17 &  79.07 \\
\modelbplarge{}              &  90.42 &  76.43 &  74.29 &  75.00 &  68.57 &  87.81 &  78.75 \\
\modelstem{}              &  89.30 &  79.29 &  70.00 &  65.00 &  75.00 &  86.70 &  77.55 \\
\modelbpmini{}               &  88.26 &  77.86 &  73.57 &  71.43 &  67.14 &  86.37 &  77.44 \\
\modelstemsuffix{}    &  90.14 &  75.71 &  \textbf{75.71} &  45.71 &  74.29 &  86.76 &  74.72 \\
\modelchar{}            &  81.41 &  73.57 &  62.14 &  68.57 &  67.86 &  82.24 &  72.63 \\
\midrule
\modellemmameta{}   &  66.20 &  67.14 &  62.14 &  67.86 &  62.14 &  67.85 &  65.56 \\
\modelstemmeta{}      &  66.01 &  47.86 &  63.57 &  62.14 &  58.57 &  67.76 &  60.99 \\
\modeltokenmeta{}        &  62.72 &  59.29 &  55.71 &  57.86 &  48.57 &  58.04 &  57.03 \\
\bottomrule
\end{tabular}
\end{threeparttable}
}
\end{table}

\subsubsection{Simplified CNN-rand Results}

\begin{table}[t!]
	\resizebox{\columnwidth}{!}{
	\centering
	\begin{threeparttable}
		\caption{Simplified CNN-rand accuracy results for sentence level sentiment analysis}
		\label{table-cnn-results} 
		\begin{tabular}{llllllll}
\toprule
{} &  Movie &   Book &    DVD & Electr. & Kitchen & Beyaz Perde & Avg \\
\midrule
\modellemmasuffix{} &  82.85 &  \textbf{72.62} &  65.44 &  72.10 &  69.78 &  73.08 &  \textbf{72.65} \\
\modelbpmid{}               &  \textbf{83.04} &  68.54 &  \textbf{66.31} &  71.89 &  \textbf{70.99} &  72.80 &  72.26 \\
\modelbplarge{}              &  82.92 &  69.13 &  66.09 &  \textbf{74.46} &  66.73 &  \textbf{73.72} &  72.18 \\
\modelstem{}              &  81.62 &  69.51 &  63.93 &  72.53 &  67.95 &  71.72 &  71.21 \\
\modellemma{}           &  82.29 &  68.54 &  64.58 &  70.39 &  68.97 &  71.76 &  71.09 \\
\modelstemsuffix{}    &  82.57 &  70.29 &  62.42 &  72.10 &  66.33 &  72.80 &  71.09 \\
\modelhybrid{}                &  79.49 &  67.57 &  63.93 &  72.75 &  67.14 &  72.97 &  70.64 \\
\modeltoken{}        &  82.92 &  69.51 &  62.20 &  67.81 &  66.94 &  73.44 &  70.47 \\
\modelsyllable{}             &  80.36 &  67.57 &  60.91 &  72.75 &  68.15 &  70.79 &  70.09 \\
\modelbpmini{}               &  78.81 &  68.54 &  62.85 &  69.31 &  67.55 &  70.50 &  69.59 \\
\modelchar{}            &  73.95 &  64.27 &  57.24 &  65.67 &  58.82 &  65.93 &  64.31 \\
\midrule
\modelstemmeta{}      &  60.95 &  63.69 &  56.59 &  61.16 &  63.08 &  58.24 &  60.62 \\
\modellemmameta{}   &  61.15 &  63.88 &  57.45 &  58.37 &  58.82 &  58.16 &  59.64 \\
\modeltokenmeta{}        &  55.85 &  57.67 &  55.29 &  53.43 &  48.68 &  53.82 &  54.12 \\
\bottomrule
\end{tabular}
\end{threeparttable}
}
\end{table}

For simplified CNN-rand, sentence level results are given in Table~\ref{table-cnn-results}. As seen in the results, \modellemmasuffix{} brings the highest average score of 72.65\%. It is followed by \modelbpmid{}, \modelbplarge{}, and \modelstem{}. For review level sentiment analysis, results are given in Table~\ref{table-cnn-results-mv}. In this case, \modellemmasuffix{} holds the highest average score of 81.27\%. It is followed by \modelbpmid{}, \modelhybrid{}, and \modelstem{}.

\begin{table}[t!]
	\resizebox{\columnwidth}{!}{
	\centering
	\begin{threeparttable}
		\caption{Simplified CNN-rand accuracy results for review level sentiment analysis}
		\label{table-cnn-results-mv} 
		\begin{tabular}{llllllll}
\toprule
{} &  Movie &   Book &    DVD & Electr. & Kitchen & Beyaz Perde & Avg \\
\midrule
\modellemmasuffix{} &  90.61 &  \textbf{81.43} &  76.43 &  77.86 &  75.00 &  86.30 &  \textbf{81.27} \\
\modelbpmid{}               &  90.61 &  75.71 &  76.43 &  75.00 &  \textbf{80.00} &  86.46 &  80.70 \\
\modelhybrid{}                &  89.01 &  75.71 &  75.00 &  \textbf{80.71} &  77.14 &  85.96 &  80.59 \\
\modelstem{}              &  88.92 &  \textbf{81.43} &  76.43 &  77.14 &  73.57 &  84.67 &  80.36 \\
\modeltoken{}        &  \textbf{91.08} &  77.14 &  75.00 &  73.57 &  77.14 &  86.81 &  80.13 \\
\modelbplarge{}              &  89.95 &  73.57 &  \textbf{77.14} &  76.43 &  76.43 &  \textbf{87.04} &  80.09 \\
\modelstemsuffix{}    &  88.83 &  77.86 &  74.29 &  80.00 &  72.14 &  85.83 &  79.82 \\
\modelsyllable{}             &  87.79 &  79.29 &  70.71 &  77.86 &  78.57 &  84.67 &  79.81 \\
\modellemma{}           &  88.92 &  76.43 &  74.29 &  73.57 &  73.57 &  85.09 &  78.64 \\
\modelbpmini{}               &  89.11 &  75.00 &  71.43 &  76.43 &  72.86 &  84.17 &  78.16 \\
\modelchar{}            &  83.76 &  69.29 &  60.71 &  71.43 &  62.14 &  79.78 &  71.18 \\
\midrule
\modellemmameta{}   &  67.23 &  72.14 &  64.29 &  57.86 &  59.29 &  66.87 &  64.61 \\
\modelstemmeta{}      &  66.20 &  72.14 &  61.43 &  54.29 &  62.14 &  65.74 &  63.66 \\
\modeltokenmeta{}        &  58.69 &  59.29 &  58.57 &  50.00 &  49.29 &  57.83 &  55.61 \\
\bottomrule
\end{tabular}
\end{threeparttable}
}
\end{table}
When we compare results for CNN-rand and Simplified CNN-rand neural network models, we can see that there is no significant difference between performances. In fact, Simplified CNN-rand accuracy results are slightly higher than original CNN-rand implementation. Average of all scores measured by the original CNN-rand implementation is 74.61\%, while the same average is 75.33\% for Simplified CNN-rand implementation. 

\subsubsection{LSTM Results}

\begin{table}[t!]
	\resizebox{\columnwidth}{!}{
	\centering
	\begin{threeparttable}
		\caption{LSTM accuracy results for sentence level sentiment analysis}
		\label{table-lstm-results} 
		\begin{tabular}{llllllll}
\toprule
{} &  Movie &   Book &    DVD & Electr. & Kitchen & Beyaz Perde & Avg \\
\midrule
\modeltoken{}        &  \textbf{83.16} &  67.57 &  \textbf{67.60} &  \textbf{73.39} &  66.73 &  73.86 &  \textbf{72.05} \\
\modelstem{}              &  81.54 &  69.51 &  66.09 &  72.75 &  69.37 &  72.63 &  71.98 \\
\modellemma{}           &  82.17 &  69.71 &  65.66 &  69.74 &  69.98 &  72.97 &  71.70 \\
\modelbpmid{}               &  80.95 &  67.77 &  \textbf{67.60} &  72.96 &  66.13 &  73.93 &  71.56 \\
\modelstemsuffix{}    &  82.25 &  69.51 &  64.36 &  71.24 &  \textbf{67.34} &  73.54 &  71.38 \\
\modelbplarge{}              &  82.81 &  67.57 &  66.09 &  68.24 &  66.94 &  \textbf{74.44} &  71.01 \\
\modellemmasuffix{} &  82.25 &  \textbf{69.90} &  66.52 &  70.39 &  62.27 &  73.81 &  70.86 \\
\modelhybrid{}                &  79.17 &  67.57 &  61.12 &  71.46 &  66.53 &  72.78 &  69.77 \\
\modelsyllable{}             &  78.93 &  69.51 &  60.69 &  71.89 &  65.52 &  71.97 &  69.75 \\
\modelbpmini{}               &  76.68 &  66.21 &  64.36 &  70.17 &  62.07 &  70.99 &  68.41 \\
\modelchar{}            &  65.42 &  60.00 &  55.72 &  58.37 &  55.38 &  67.13 &  60.33 \\
\midrule
\modelstemmeta{}      &  59.25 &  61.17 &  57.24 &  60.30 &  54.77 &  58.87 &  58.60 \\
\modellemmameta{}   &  58.77 &  60.97 &  56.59 &  58.15 &  53.75 &  58.40 &  57.77 \\
\modeltokenmeta{}        &  55.57 &  57.48 &  55.94 &  46.35 &  53.75 &  54.55 &  53.94 \\
\bottomrule
\end{tabular}
\end{threeparttable}
}
\end{table}

Sentence level results for LSTM are given in Table~\ref{table-lstm-results}. As given in the table, \modeltoken{} provides the highest average score of 72.05\%. It is followed by \modelstem{}, \modellemma{}, \modelbpmid{}, and \modellemmasuffix{} with similar accuracy results, respectively. Review level results for LSTM are presented in Table~\ref{table-lstm-results-mv}. According to the results, \modeltoken{} gives the highest average score of 81.36\%. It is followed by \modelbpmid{}, and \modellemma{} with similar accuracy performance.

\begin{table}[t!]
	\resizebox{\columnwidth}{!}{
	\centering
	\begin{threeparttable}
		\caption{LSTM accuracy results for review level sentiment analysis}
		\label{table-lstm-results-mv} 
		\begin{tabular}{llllllll}
\toprule
{} &  Movie &   Book &    DVD & Electr. & Kitchen & Beyaz Perde & Avg \\
\midrule
\modeltoken{}        &  \textbf{90.80} &  77.86 &  76.43 &  \textbf{80.71} &  75.00 &  87.39 &  \textbf{81.36} \\
\modelbpmid{}               &  89.86 &  75.71 &  75.71 &  79.29 &  75.00 &  \textbf{87.91} &  80.58 \\
\modellemma{}           &  89.20 &  80.00 &  77.86 &  75.00 &  75.00 &  86.20 &  80.54 \\
\modelbplarge{}              &  89.30 &  72.14 &  \textbf{79.29} &  74.29 &  \textbf{77.14} &  88.56 &  80.12 \\
\modelstem{}              &  89.39 &  78.57 &  78.57 &  78.57 &  68.57 &  86.28 &  79.99 \\
\modelhybrid{}                &  89.48 &  75.00 &  73.57 &  80.00 &  74.29 &  86.67 &  79.83 \\
\modelstemsuffix{}    &  90.05 &  80.71 &  74.29 &  76.43 &  70.71 &  86.80 &  79.83 \\
\modellemmasuffix{} &  90.33 &  77.86 &  \textbf{79.29} &  77.14 &  65.71 &  87.06 &  79.56 \\
\modelsyllable{}             &  87.23 &  \textbf{82.14} &  70.71 &  75.00 &  75.00 &  85.61 &  79.28 \\
\modelbpmini{}               &  86.48 &  70.00 &  72.14 &  73.57 &  67.14 &  85.43 &  75.79 \\
\modelchar{}            &  72.58 &  60.71 &  55.00 &  64.29 &  60.00 &  81.37 &  65.66 \\
\midrule
\modelstemmeta{}      &  65.26 &  65.71 &  64.29 &  65.00 &  57.86 &  67.43 &  64.26 \\
\modellemmameta{}   &  64.23 &  68.57 &  58.57 &  62.86 &  61.43 &  66.00 &  63.61 \\
\modeltokenmeta{}        &  57.37 &  58.57 &  64.29 &  45.71 &  52.86 &  57.94 &  56.12 \\
\bottomrule
\end{tabular}
\end{threeparttable}
}
\end{table}

\subsubsection{Results in Comparison to Baseline Methods}

In Table~\ref{table-results-beyazperde}, we present our results on \datasetbeyazperde{} in comparison to two previous studies in the literature, SentiStrength in~\citet{Vural2012} and Feature Selection in \citet{uccan2014otomatik}. The table shows that the best performing segmentation methods on CNN and LSTM  outperform the best results in baseline scores, which belong to~\citet{uccan2014otomatik}. On average our scores are $+3$ points higher than results in~\citet{uccan2014otomatik}. LSTM @~\modelbpmid{}, outperforms it by a large $+4$ point margin.

\begin{table}[t!]
\centering
\begin{threeparttable}
\caption{Accuracy Results on the \datasetbeyazperde{} data set}
\label{table-results-beyazperde} 
\begin{tabular}{llc}
\toprule
\textbf{Model}		 & Author & \textbf{Acc.}\\
\midrule
Sentence-binary	 & \citet{Vural2012}	& 70.39 \\ 
Sentence-max/min	 & \citet{Vural2012}	& 74.83 \\ 
Word-sum 			 & \citet{Vural2012}	& 75.90 \\ 
Chi Square			 & \citet{uccan2014otomatik}	& \textbf{83.90} \\ 
Information Gain	 & \citet{uccan2014otomatik}	& \textbf{83.90} \\ 
\midrule
CNN @~\modellemmasuffix{} & 	 & 86.30 \\
CNN @~\modelbpmid{}      & 	 & 86.46 \\
CNN @~\modelhybrid{}     & 	 & 85.96 \\
LSTM @~\modeltoken{}     & 	 &  87.39 \\
LSTM @~\modelbpmid{}     & 	 &  \textbf{87.91} \\
LSTM @~\modellemma{}     & 	 &  86.20 \\
\bottomrule 
\end{tabular}
\end{threeparttable}
\end{table}

\begin{table}[t!]
\centering
\begin{threeparttable}
\caption{Accuracy results on the \datasetproducts{} data set}
\label{table-results-products} 
\begin{tabular}{lccccc}
\toprule
\textbf{Model} & \textbf{Movie} & \textbf{Book} & \textbf{DVD} & \textbf{Electronics} & \textbf{Kitchen}\\
\midrule
Naive Bayes$^1$ 		 & 69.50 & \textbf{72.40} & \textbf{76.00} & \textbf{73.00} & \textbf{75.90}  \\ 
Naive Bayes MT$^1$ 		 & \textbf{70.00} 	 & 71.70 	 & 74.90 	 & 64.40 	 & 69.60  \\ 
Linear SVC$^1$ 			 & 66.00 	 & 66.60 	 & 70.30 	 & 72.40 	 & 70.00  \\ 
Linear SVC MT$^1$ 		 & 66.50 	 & 66.90 	 & 67.60 	 & 64.40 	 & 67.30  \\ 
\midrule
CNN @~\modellemmasuffix{} &  90.61 &  \textbf{81.43} &  76.43 &  77.86 &  75.00 \\
CNN @~\modelbpmid{}               &  90.61 &  75.71 &  76.43 &  75.00 &  \textbf{80.00} \\
CNN @~\modelhybrid{}                &  89.01 &  75.71 &  75.00 &  \textbf{80.71} &  77.14 \\
LSTM @~\modeltoken{}        &  \textbf{90.80} &  77.86 &  76.43 &  \textbf{80.71} &  75.00 \\
LSTM @~\modelbpmid{}               &  89.86 &  75.71 &  75.71 &  79.29 &  75.00 \\
LSTM @~\modellemma{}           &  89.20 &  80.00 &  \textbf{77.86} &  75.00 &  75.00 \\
\bottomrule 
\end{tabular}
\begin{tablenotes}
	\small
	\item $^1$\citet{Demirtas_2013}
\end{tablenotes}
\end{threeparttable}
\end{table}

We also present our results on \datasetproducts{} in Table~\ref{table-results-products} in comparison to earlier scores derived from the dataset by~\citet{Demirtas_2013}. Our results outperform baseline scores in all categories. In average, the margin between our results and best baseline scores is $6.7$.

\begin{table}[t!]
\centering
\caption{Margin between best baseline scores and our results on \datasetbeyazperde{}, and \datasetproducts{} data sets}
\label{table-results-margin} 
\begin{tabular}{lccccccc}
\toprule
{} & Movie & Book & DVD & Electr. & Kitchen & Beyaz Perde & \textbf{Avg}\\
\midrule
Overall 				 &  \textbf{+21.1} &  \textbf{+9.8} &  \textbf{+3.3} &  \textbf{+7.7\-} &  \textbf{+4.1} & \textbf{+4.0} & \textbf{+8.3} \\
Selected Models 	     &  +20.8 &  +9.0 &  +1.9 &  \textbf{+7.7\-} &  \textbf{+4.1} & \textbf{+4.0} & +7.9 \\
CNN @~\modelbpmid{}      &  +20.6 &  +3.3 &  +0.4 &  +2.0 &  \textbf{+4.1} & +2.6 & +5.5 \\
\bottomrule 
\end{tabular}
\end{table}

In fact, we have a single model that outperforms all baseline scores alone, which is CNN @~\modelbpmid{}. This segmentation method is among  top 3 results in both CNN and LSTM neural models, holding the second position in both. In Table~\ref{table-results-margin} we present overall margins between results of our study and the best baseline scores. The table shows that CNN @~\modelbpmid{} classifier outperforms baseline scores on average of $+5.5$ points. Consolidated results from selected classifiers outperform baseline scores on an average of $+7.92$ points, and overall results on an average of $+8.21$ points.

\subsection{Performance Evaluation}
In this part, we will evaluate performances of segmentation methods, and neural networks and investigate into what they do good and what they could do better. 

\subsubsection{Distribution of Polarity Predictions and Majority Voting}

Sentiment extraction for positive or negative text classification is a binary problem. Therefore, neural networks are trained to output a value between 0-1 according to how certain the model is about the polarity of the input text. As a rule of thumb, values near 0.5 are those uncertain and they are very likely to be neutral. Similarly, a prediction near 1 or 0 is one the model is pretty sure that it is negative or positive. For instance, a prediction value 0.96 will indicate that model thinks the input text is almost certainly positive. 

In order to evaluate performances of models, we will first compile entire predictions into one single prediction file. From there, we can extract distribution of a particular neural network with a particular segmentation method. Since only \modelbpmid{} was able to take its place among selected segmentation methods both for CNN and LSTM, and also because CNN @~\modelbpmid{} outperforms every baseline score, we will use \modelbpmid{} as our benchmark segmentation model. However, we will occasionally switch from CNN to LSTM when needed. 

\begin{figure}[t!]
	\centering
	\includegraphics[width=12cm]{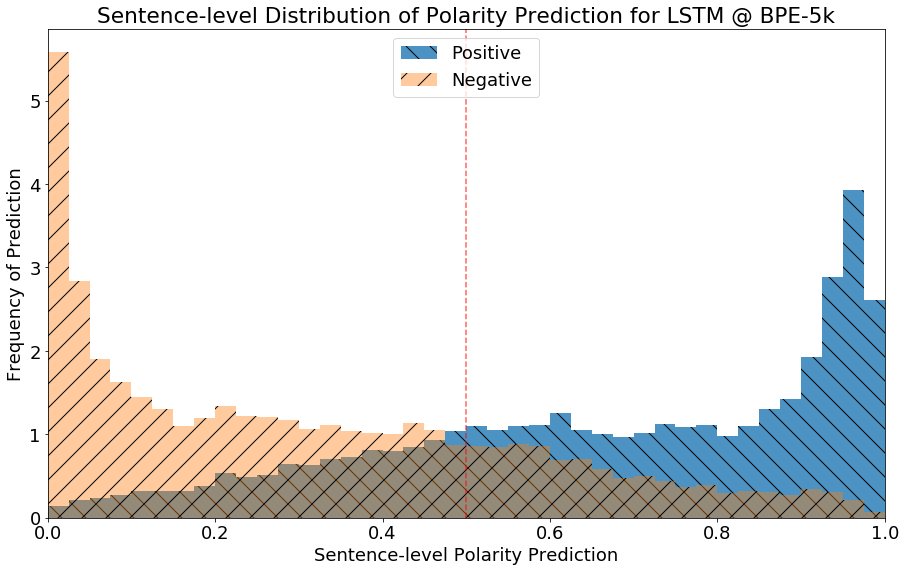}
	\caption{Predicted polarity distribution of test sample data for an LSTM networks trained with \modelbpmid{} segmentation output. Distribution is Sentence-level.}
	\label{lstm-bpe-5k-sentence-level-polarity-prediction-histogram}
\end{figure}

Figure~\ref{lstm-bpe-5k-sentence-level-polarity-prediction-histogram} shows the distribution of polarity prediction by our LSTM network trained by \modelbpmid{} segmentation output. Orange area shows the distribution of prediction for sentences labeled as Negative, and blue area shows the distribution of sentences labeled as Positive. One important aspect in this figure is that it is similar to a Chi-squared distribution with $k=2$. However, there are too many values in between, possibly due to data not being clean enough. Hence, this is overall an expected distribution. Our model is specifically trained to output binary values for classification.

The overlapping areas constitute samples classified incorrectly. From average accuracy score of \modelbpmid{}, we can deduct the ratio of the overlapping area. Table~\ref{table-lstm-results} indicates that the average score for \modelbpmid{} is 71.56, which means that the overlapping area constitutes $100-71.56 = 28.44\%$ of overall distribution.

\begin{figure}[t!]
	\centering
	\label{lstm-bpe-5k-review-level-polarity-prediction-histogram}
	\caption{Predicted polarity distribution of test sample data for an LSTM networks trained with \modelbpmid{} segmentation output. Distribution is Review-level}
	\includegraphics[width=12cm]{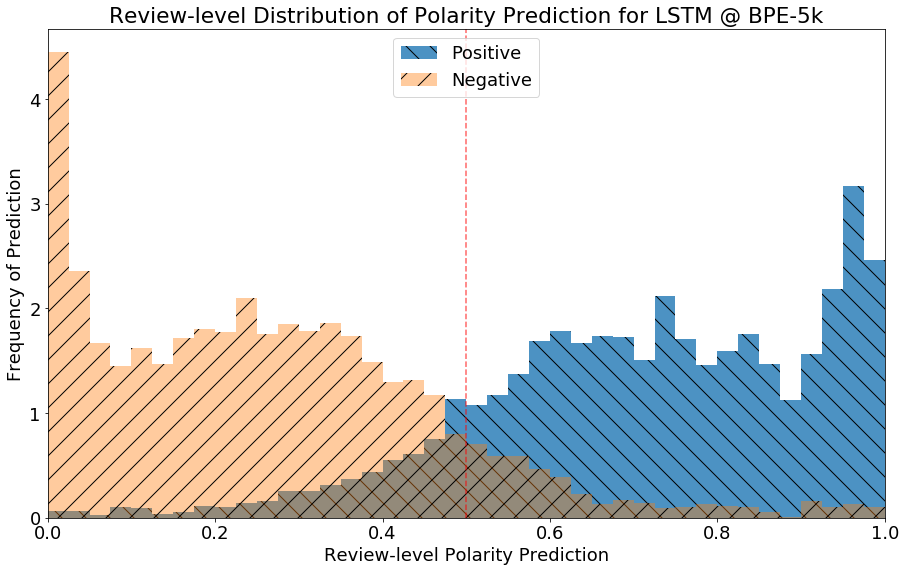}
\end{figure}

In Figure~\ref{lstm-bpe-5k-review-level-polarity-prediction-histogram} we can see the distribution of review-level polarity prediction frequency distribution. In this case, the distribution is normal in comparison to the initial distribution. This is also an expected behavior if we take into account how the majority voting process works. 

During pre-processing we divided reviews into sentences only to be compiled into the same review after polarity predictions are derived from neural network classifier. The compilation works in a way that a small sample of sentences is taken from sentence sample and the combined average predicted value is calculated for the review. The procedure we are describing here is the one that \emph{Central Limit Theorem} (CLT) adheres to explain. CLT indicates that, when a sample size of n is drawn repeatedly from a distribution, the distribution of means of these samples will have a standard deviation ($\sigma{}$) value of square root times the $\sigma{}$ of the original distribution. On the other hand, It will still have the same mean value of$\mu$. See Equation~\ref{clt-std-formula} for reference.

\begin{align}
\label{clt-std-formula}
\begin{aligned}
\mathllap{\sigma{}_n} = \frac{\sigma{}_0}{\sqrt{n}}
\end{aligned}
\end{align}

This means that the new distribution will be converging to a normal distribution with smaller standard deviation around the same mean. Majority voting creates a new distribution in the same way, and therefore, expected to have a normal distribution around mean value of the original distribution.

\begin{figure}[t!]
	\centering
	\includegraphics[width=12cm]{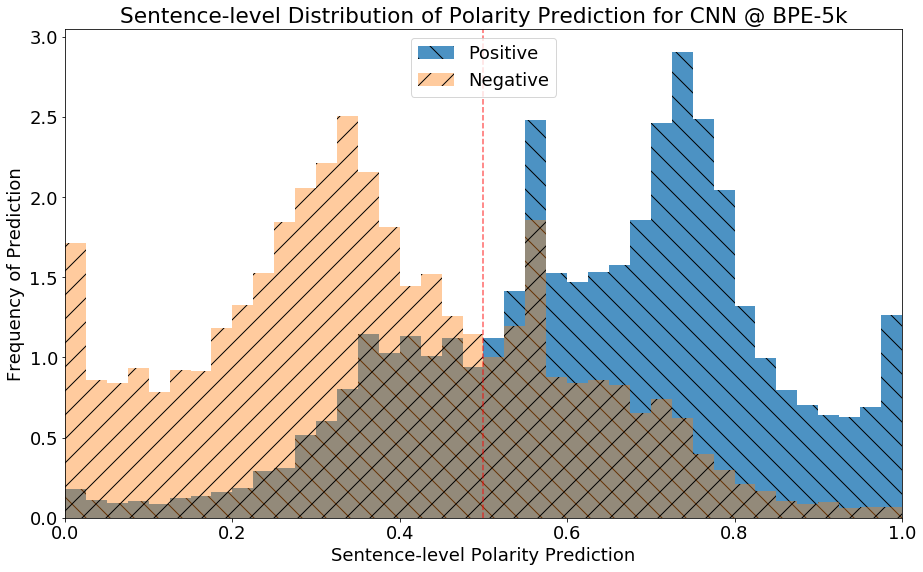}
	\caption{Predicted polarity distribution of test sample data for a CNN networks trained with \modelbpmid{} segmentation output. Distribution is Sentence-level}
	\label{cnn-bpe-5k-sentence-level-polarity-prediction-histogram}
\end{figure}

Figure~\ref{cnn-bpe-5k-sentence-level-polarity-prediction-histogram} shows the distribution of polarity prediction by our CNN network trained by \modelbpmid{} segmentation output. This distribution is quite different than the LSTM sentence-level distribution.  What is quite surprising about it is the fact that distribution is a normal distribution with an excess accumulation at the outer edges. This is strange because in a way it means the CNN network is not very certain about the polarity of the majority of test entries fed into the model. The overlapping area is also accumulated around the neutral region. This area is more spread-out in LSTM distribution. 

\begin{figure}[t!]
	\centering
	\includegraphics[width=12cm]{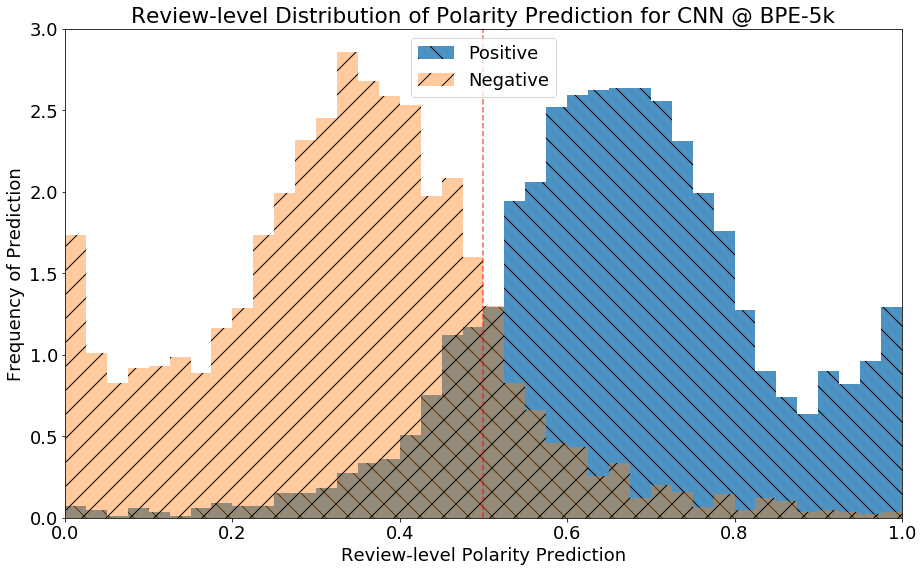}
	\caption{Predicted polarity distribution of test sample data for a CNN networks trained with \modelbpmid{} segmentation output. Distribution is Review-level}
	\label{cnn-bpe-5k-review-level-polarity-prediction-histogram}
\end{figure}

In Figure~\ref{cnn-bpe-5k-review-level-polarity-prediction-histogram} we can see the distribution of review-level polarity prediction frequency distribution for CNN network. Due to the same reasons that we listed for LSTM review-level distribution, the new sample space will have a more accumulated normal distribution thanks to Central Limit Theorem.

\subsubsection{Performance Comparison on Sample Cases}

\begin{table}[t!]
	\centering
	\begin{threeparttable}
		\caption{Sample Predictions for a review compiled from LSTM @~\modelbpmid{} and LSTM @~\modellemmasuffix{}}
		\label{sample-prediction-output} 
\begin{tabular}{lrrr}
	\toprule
	Sentence &  Label & C1$^1$ & C2$^2$ \\
	\midrule
	\midrule
\pbox{10cm}{elif şafakın imzalı kitabını hepsiburadan temin etmek çok güzel . \newline(being able to acquire elif şafak's[typo] signed book from hepsiburada[typo] is very nice .)} &  1.0 &  0.36 &  0.83 \\
\midrule
\pbox{10cm}{bu fırsat kaçırılmamalı . \newline(this opportunity shouldn't be missed .)} &  1.0 &  0.23 &  0.79 \\
\midrule
\pbox{10cm}{hediye olarak çok güzel bir kitap . \newline(this is a very good book for a present .)} &  1.0 &  0.77 &  0.88 \\
\midrule
\pbox{10cm}{konudan bahsetmeyeceğim , elif şafakı tanıyanlar bilir ... \newline(i won't talk about the theme , those that know elif şafak(typo) will know ...)} &  1.0 &  0.78 &  0.21 \\
	\bottomrule
\end{tabular}

\begin{tablenotes}
\small
\item $^1$LSTM @~\modelbpmid{}
\item $^2$LSTM @~\modellemmasuffix{}
\end{tablenotes}
\end{threeparttable}
\end{table}

Table~\ref{sample-prediction-output} shows a sample prediction for a random review extracted from predictions derived from LSTM @~\modelbpmid{} and LSTM @~\modellemmasuffix{}. The review constitutes 4 different sentences. The LSTM @~\modelbpmid{} classifier provided 0.36, 0.23, 0.76, and 0.78 scores respectively for each sentence. The mean value is 0.53, and larger than 0.5. Therefore, we can conclude that the review is classified correctly by a small margin. The LSTM @~\modellemmasuffix{} classifier, however, provides much better scores than the former. It derives 0.83, 0.79, 0.88, and 0.21 scores for sentences respectively. The average is 0.67, and unlike \modelbpmid{} the review is classified as positive by a larger margin.

From a manual inspection, we can conclude that the review is clearly positive, and the author or review is praising the author of the book s/he is reviewing. The distinction is important in a way that it can distinguish what a morphological analysis such as \modellemmasuffix{} can achieve better than a sub-word model such as \modelbpmid. However, we also know that \modelbpmid{} has a better overall performance than \modellemmasuffix. Therefore we can conclude that there are many factors at play, and the overall schema opts for the \modelbpmid.

\begin{table}[!h]
	\centering
	\begin{threeparttable}
		\caption{Sample Predictions for a review compiled from CNN @~\modelbpmid{} and CNN @~\modellemmasuffix{}}
		\label{sample-prediction-output-cnn} 
\begin{tabular}{lrrr}
	\toprule
	Sentence &  Label & C1$^1$ & C2$^2$ \\
	\midrule
	\midrule
\pbox{10cm}{quantin tarantino iyi bir yönetmen ama ben bu kadar kötü bir film beklemezdim .\newline(quantin tarantino is a good director, but i wouldn't expect a movie this bad .)} &  0.0 &  0.23 &  0.24\\
\midrule
\pbox{10cm}{film çok sıkıcı başladı inatla sonuna kadar izledim acaba değişir mi diye ama hüsran .\newline(movie started dull, i resisted to the end hoping it would change, but disappointment .)} &  0.0 &  0.11 &  0.04\\
\midrule
\pbox{10cm}{bana göre gerçekten kötü bir film \newline(to me it is a really bad movie)} &  0.0 &  0.29 &  0.15\\
	\bottomrule
\end{tabular}

\begin{tablenotes}
\small
\item $^1$CNN @~\modelbpmid{}
\item $^2$CNN @~\modellemmasuffix{}
\end{tablenotes}
\end{threeparttable}
\end{table}

Table~\ref{sample-prediction-output-cnn} shows sentiment predictions by CNN with \modelbpmid{} and \modellemmasuffix{} segmentations models for a sample review. Table shows that the negative review has an average sentiment score of 0.21 and 0.14 for \modelbpmid{} and \modellemmasuffix{} respectively. We can once again conclude that even if \modelbpmid{} can be more robust than most other candidates in terms of accuracy, the sentiment prediction is more certain with \modellemmasuffix{}, and thereby with morphological approaches.

\subsubsection{Cross-parameter Comparison}

In order to understand the relationship between different parameters and their effect on accuracy results, we compiled various parameters for performance evaluation during model training. 

During our experiments, we collected a wide range of measurements on preprocessing, training, and evaluation. These results are available both for CNN and LSTM experiments. The parameters measured and collected are shown in Table~\ref{list-of-experiment-measurements} with their descriptions.

\begin{table}[t!]
	\centering
	\begin{threeparttable}
		\caption{List of parameters collected during experiments with pre-precessing and neural model building.}
		\label{list-of-experiment-measurements} 
\begin{tabular}{ll}
	\toprule
	Parameter &  Description \\
	\midrule
	\midrule
 	No &  The order in which the experiment is executed.\\
 	\midrule
	Dataset & Name of dataset \\
	\midrule
	Segmentation &  Segmentation method used during preprocessing \\
	\midrule
	Train & Size of training sample \\
	\midrule
	Validation & Size of validation sample \\
	\midrule
	Test & Size of test sample \\
	\midrule
	Batch Size & Batch size used model training$^*$\\
	\midrule
	Vocabulary & Vocabulary size \\
	\midrule
	Max Review Length & \pbox{9cm}{Size of longest review in dataset after pre-processing} \\
	\midrule
	Pre-processing Duration & Pre-processing duration in seconds \\
	\midrule
	Train Duration & Model training duration in seconds \\
	\midrule
	Evaluation Duration & Test data prediction duration \\
	\midrule
	Score & Sentence-level raw score \\
	\midrule
	MV Score & Review-level score after Majority Voting \\
	\midrule
	Epoch Count & \pbox{9cm}{Number of epochs run until early stopping kicks in} \\
	\midrule
	Save Epoch & \pbox{9cm}{The epoch number the best model is encountered and saved} \\
	\bottomrule
\end{tabular}

\begin{tablenotes}
\small
\item $^*$ Batch size is calculated after preprocessing. The final batch size depends on vocabulary and maximum review length.
\end{tablenotes}
\end{threeparttable}
\end{table}

\begin{figure}[t!]
	\centering
	\includegraphics[width=12cm]{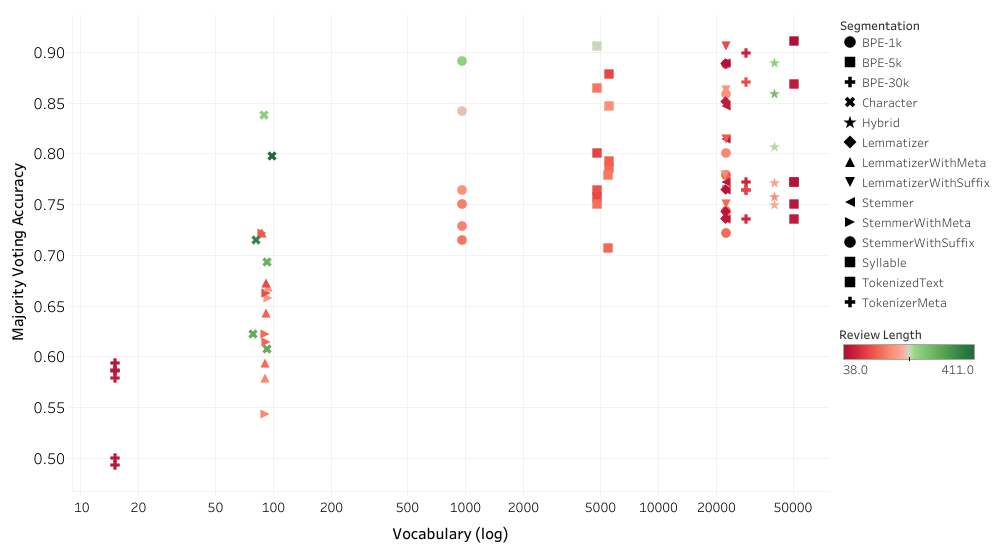}
	\caption{Scatter plot showing relation between vocabulary size and accuracy among all results acquired with CNN. Review length and Segmentation are added as details.}
	\label{accuracy-vocabulary-segmentation-comparison}
\end{figure}

In Figure~\ref{accuracy-vocabulary-segmentation-comparison} we can see the relationship between majority voting accuracy, and vocabulary. The details Segmentation and Review length parameters add further insight. We can see that there is a clear correlation between vocabulary size and accuracy. However, the relation is possibly non-linear, and the accuracy will start to deteriorate after a certain vocabulary size. In fact, our most promising segmentation \modelbpmid{} has a vocabulary size of 5k. This figure itself also hints at deteriorating nature of this correlation around 20k vocabulary size. However, further research will be needed to prove this claim. We will not go into further detail in this work.

\subsubsection{Computational Performance}

In this part, we will address three fundamental performance metrics. These metrics are of a high value in order to express a definitive way of dealing with sentiment analysis task in Turkish texts on a service level. An efficient way of dealing with this task is creating the most value with the least investment. Therefore, we will be investigating the performance of each classifier alongside the resources they use during sentiment extraction.

In order to determine the resources model use, we identified a set of metrics, and kept measurements regarding these metrics. In order to measure memory allocation, we calculated the total memory neural network needs in order to run. Since neural networks built by Keras running on Tensorflow back-end are able to use Graphics Processing Unit (GPU) the memory allocation here will represent GPU memory used for the neural network. On the other hand, we also determined the parameters affecting the total memory neural networks allocate to run. Formulas~\ref{new-parameters-1} and \ref{new-parameters-2} show how to calculate total memory a CNN or an LSTM network allocate based on parameters such as vocabulary and maximum input size. 

On the other hand, in order to calculate processing power allocated to the classifiers we identified the duration of calculations as the candidate measurement. Our implementation uses three processing steps in conjunction with each other in order to pre-process data, train a model based on this, and test the model with the test subset. The code designed to do this also keeps track of durations all these steps take. The first time-stamp ($t_0$) is recorded during the initialization, where the dataset is to be pushed into data pre-processing unit. The second time-stamp ($t_1$) is recorded when the pre-processing is complete. The difference between the two ($t_1 - t_0$) provides pre-processing duration. After this point, model building and training starts. The third time-stamp ($t_2$) is recorded after training stage is complete. The difference between third and second time-stamps ($t_2 - t_1$) gives the model training duration. Finally, the test data subset is fed into the trained model for sentiment predictions. When prediction stage is complete, the last time-stamp ($t_3$) is recorded. The difference between the last and third time-stamps ($t_3 - t_2$) gives the evaluation duration. 

The first and the most important metric is the accuracy, which we covered throughout the report. Without a plausible accuracy rate, a faster classifier with a smaller footprint will be just as meaningless. We must first explore and discover an efficient way of achieving a good accuracy in order to go on with computational performance related issues.

The second one is how much resources it takes up during training and evaluation. This is also crucial if we are going to deploy the implementation and neural modal for regular usage. The resources could be considered in two main groups, memory and computation. 

The third one is how fast it is with extracting sentiment from a sample input text. 

The First performance metric is heavily addressed in earlier parts, therefore, we will only add it to discussions as part of investigation detail. Resource consumptions and prediction speed will be the primary topics. We will use data described in Table~\ref{list-of-experiment-measurements} for performance evaluations in this part.

Not all parameters we need exists within our data, some of them will be calculated based on available parameters. We will create parameters for memory consumed by CNN and LSTM networks, and training duration until the best model is encountered.
 
The new parameters can be defined as follows.

\begin{align}
&\begin{aligned}
\label{new-parameters-1}
\mathllap{M_{CNN}} &= 50*v+500*l+3121
\end{aligned}\\
&\begin{aligned}
\label{new-parameters-2}
\mathllap{M_{LSTM}} &= 32*v+0*l+53301
\end{aligned}\\
&\begin{aligned}
\label{new-parameters-3}
\mathllap{t_{atd}} &=t_{training}*(1-(EC-SE)/EC)
\end{aligned}\\
&\begin{aligned}
\label{new-parameters-4}
\mathllap{t_{te}} &=t_{pp}+t_{eval}
\end{aligned}
\end{align}

Where;
\begin{description}
	\item[$v$] denotes \emph{Vocabulary size for preprocessed dataset},
	\item[$l$] denotes \emph{Max Review Length for neural network model input},
	\item[$M_{CNN}$] denotes \emph{Amount of Memory the CNN model uses},
	\item[$M_{LSTM}$] denotes \emph{Amount of memory the LSTM model uses},
	\item[$t_{training}$] denotes \emph{The duration needed to train the network},
	\item[$EC$] denotes \emph{Total amount of Epochs until the training is finalized},
	\item[$SE$] denotes \emph{The epoch on which the best results encountered during training},
	\item[$t_{atd}$] denotes \emph{Actual Train Duration: Calculated training duration until best results are encountered},
	\item[$t_{pp}$] denotes \emph{Pre-processing Duration: The time it takes to process raw input into the shape the network can process},
	\item[$t_{eval}$] denotes \emph{Evaluation Duration: The time it takes to evaluate a preprocessed input},
	\item[$t_{te}$] denotes \emph{Total Evaluation Duration: Total amount of time it take to pre-process raw data and evaluate it}.
\end{description}
 
 \begin{figure}[t!]
	\centering
	\includegraphics[width=12cm]{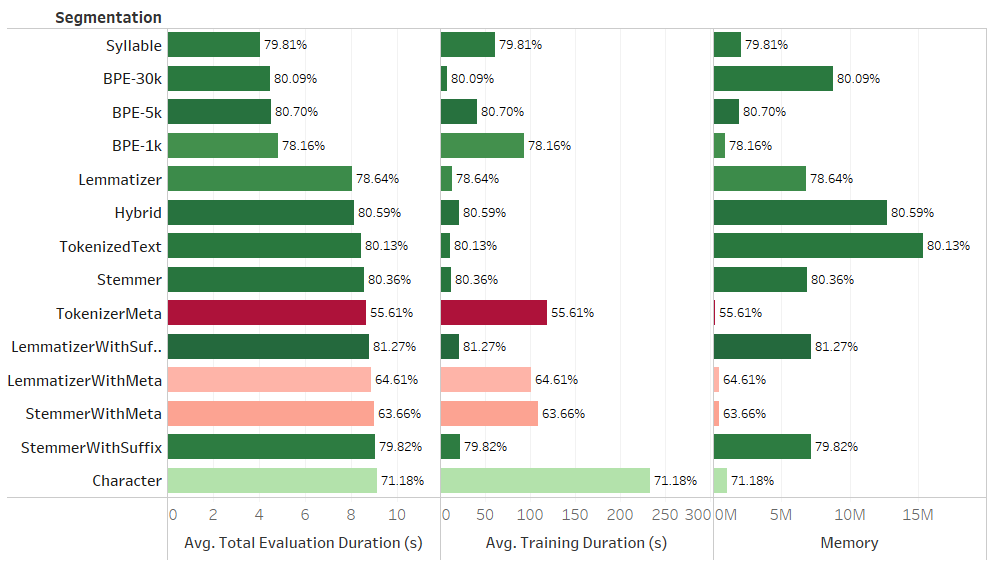}
	\caption{Bar chart showing average memory size, average test data prediction duration, and model memory for a CNN network.}
	\label{cnn-memory-preprocessing-duration}
\end{figure}

We extracted the new parameters by using formulas provided in ~\ref{new-parameters-1},~\ref{new-parameters-2},~\ref{new-parameters-3}, and \ref{new-parameters-4}. Using new parameters we plotted the relationships in column charts both for Evaluation Duration and Training Duration. Figure~\ref{cnn-memory-preprocessing-duration} shows the distribution of these two parameters among different segmentation methods. The columns are sorted by ascending average evaluation duration. 

From the figure we can deduct that \modelsyllable{}, \bpelist{} methods provide shortest total evaluation durations. Note that \emph{Total Evaluation Duration} is the sum of test data pre-processing, and prediction durations. On the other hand, they also hold some of highest scores for accuracies. \modelbpmid{} seems to be an obvious candidate for the most efficient model in terms of accuracy, training duration, and memory.

\subsection{Summary}

The results we acquired during experiments show that there is substantial support for nominating \modelbpmid{} segmentation method as the most efficient one combining our metrics both for accuracy and computational performance. However, one important thing to point out is that the difference between the accuracy results among prominent segmentation methods are not very strong. There is also strong evidence that the performances of CNN and LSTM networks seem to be similar. 

The following points can be given as the summary of the results presented in this work.

\begin{itemize}
	\item \modelbpmid{} performs good both in terms of accuracy and computational performance.
	\item CNN and LSTM networks perform similarly. 
	\item Widely used \modeltoken{} segmentation method performs as good as other ones with similar scores in terms of accuracy.
	\item The positional attributes of words in sentences without word themselves keeps a substantial amount of info by which these models achieved accuracy results around 65\%. This also means that this information could be used as an additional layer in future.
\end{itemize}

\section{Conclusion and Future Work}
\label{chapter:fw}

Our results clearly show that neural network models are more successful as carefully engineered lexicon and rule-based methods, mostly by a large margin. 
\modellemmasuffix, \modelbpmid, \modelhybrid, \modeltoken, and \modellemma{} proven to be the best ones. Our selected models outperformed best scores in baselines by an average $+7.92$ points. 
We also noted \modelbpmid{} model for its performance with being among top 3 for results acquired by using both CNN and LSTM based classifiers. It also achieves better scores than any baseline score alone. This model outperformed best scores of baselines by an average $+5.5$ points along all datasets. 

Our prediction distributions showed that LSTM networks seem to be more certain about their predictions with a Chi-squared distribution. (Figure~\ref{lstm-bpe-5k-sentence-level-polarity-prediction-histogram}) 
Whereas, CNN distribution is normal near neutral region. (Figure~\ref{cnn-bpe-5k-sentence-level-polarity-prediction-histogram})
We also observed Central Limit Theorem being at play with our Majority Voting process. We also validated the effect with CLT formulas. (Equation~\ref{clt-std-formula})
We reviewed some sample outputs and their predictions by different segmentation methods. We observed that even if they predict the same output, they do it for seemingly quite different reasons, which showed that there are very complex factors being at play for different segmentation methods. (Table~\ref{sample-prediction-output})
We also investigated the relationships between model scores, segmentation methods and review lengths. We observed that the networks performed better until a certain vocabulary size.  (Figure~\ref{accuracy-vocabulary-segmentation-comparison})

This work did not focus on advanced natural language processing techniques such as phrase and idiom extraction, and sentence attention detection. It is not aimed to extract sentence forms and types, as well. We believe that, by employing such methods, scores acquired in this work could be improved. In addition, by inspecting the predictions which provide poor accuracy the more important features could be determined. In this work, we also did not try to check and correct typographical errors within the text. Informal texts are known to be hosting too many errors of this type. By employing spell-checking and advanced error correction mechanisms vocabulary could be reduced and accuracies could be improved. In addition, text can be normalized where information is formulated or represented in other forms such as numbers. Normalization might both reduce vocabulary size and increase accuracy. We think that using typo checkers and text normalization can add further improvement to our models and accuracy scores.
We also did not intend to use advanced disambiguation for words while extraction positional attributes and suffixes. Turkish has too many seemingly similar or the same, but structurally different words due to its rich morphology. Developing and utilizing a tool capable of understanding the context of the sentence and determining the most appropriate variant of these words could improve accuracy performances our models further. For instance, the word \emph{kara} (dark; land; into snow (in eng.)) is used with different meanings in following phrases.

\begin{itemize}
	\item \textbf{Kara} bulutlar gökleri doldurdu. (\textbf{Dark} clouds filled the skies.)
	\item Ada Marmarada'ki ufak bir \textbf{kara} parçasından oluşmaktadır. (The island is a small \textbf{land} in Marmara.)
	\item \textbf{Kara} batan ayaklarını hızlıca çekti. (He quickly pulled his legs which were sinking \textbf{into snow}.)
\end{itemize}

As another improvement, LSTM networks that are capable of processing multiple layers of data can be used. 
By this way, various segmentation methods presented in this work could be assigned as different layers to a new multi-layered LSTM network. We believe such an arrangement could improve the performance of the sentiment analysis task. Finally, we did not use \emph{word2vec} libraries in order to build embeddings for neural models. The most important reason of this choice is that our segmentation methods divide each word into sub-word fragments which by themselves mostly do not have a meaning or a meaning related to the word used. However, it is still possible to build a word2vec for each segmentation method by calculating weights for each of these fragments by building a model that will calculate the ratio of each fragment in various words.

\starttwocolumn
\bibliography{references}

\begin{thebibliography}{43}
\expandafter\ifx\csname natexlab\endcsname\relax\def\natexlab#1{#1}\fi

\bibitem[{Ahmet Afsin~Akin(2007)}]{Akin2007}
Ahmet Afsin~Akin, Mehmet Dundar~Akin. 2007.
\newblock Zemberek, an open source nlp framework for turkic languages.

\bibitem[{A{\c s}lıyan and G{\"u}nel(2007)}]{asliyan2007}
A{\c s}lıyan, Rıfat and Korhan G{\"u}nel. 2007.
\newblock Design and implementation for extracting turkish syllables and
  analyzing turkish syllables.

\bibitem[{Baccianella, Esuli, and
  Sebastiani(2010)}]{Baccianella10sentiwordnet3.0}
Baccianella, Stefano, Andrea Esuli, and Fabrizio Sebastiani. 2010.
\newblock Sentiwordnet 3.0: An enhanced lexical resource for sentiment analysis
  and opinion mining.
\newblock In \emph{in Proc. of LREC}.

\bibitem[{Boynukalin(2012)}]{Boynukalin2012}
Boynukalin, Zeynep. 2012.
\newblock Emotion analysis of turkish texts by using machine learning methods.
\newblock Master's thesis, Middle East Technical University.

\bibitem[{Cambria, Olsher, and
  Rajagopal(2014)}]{Cambria:2014:SCC:2892753.2892763}
Cambria, Erik, Daniel Olsher, and Dheeraj Rajagopal. 2014.
\newblock Senticnet 3: A common and common-sense knowledge base for
  cognition-driven sentiment analysis.
\newblock In \emph{Proceedings of the Twenty-Eighth AAAI Conference on
  Artificial Intelligence}, AAAI'14, pages 1515--1521, AAAI Press.

\bibitem[{Coban, Ozyer, and Ozyer(2015)}]{Coban_2015}
Coban, Onder, Baris Ozyer, and Gulsah~Tumuklu Ozyer. 2015.
\newblock Sentiment analysis for turkish twitter feeds.
\newblock In \emph{2015 23nd Signal Processing and Communications Applications
  Conference ({SIU})}, Institute of Electrical and Electronics Engineers
  ({IEEE}).

\bibitem[{{\c{C}}{\"o}ltekin(2014)}]{L14-1375}
{\c{C}}{\"o}ltekin, {\c{C}}a{\u{g}}r{\i}. 2014.
\newblock A set of open source tools for turkish natural language processing.
\newblock In \emph{Proceedings of the Ninth International Conference on
  Language Resources and Evaluation (LREC-2014)}, European Language Resources
  Association (ELRA).

\bibitem[{{\c{C}}{\"o}ltekin, Boz{\c{s}}ahin
  et~al.(2007)}]{ccoltekin2007syllables}
{\c{C}}{\"o}ltekin, {\c{C}}a{\u{g}}r{\i}, Cem Boz{\c{s}}ahin, et~al. 2007.
\newblock Syllables, morphemes and bayesian computational models of acquiring a
  word grammar.
\newblock In \emph{Proceedings of the Annual Meeting of the Cognitive Science
  Society}, volume~29.

\bibitem[{Dehkharghani et~al.(2016{\natexlab{a}})Dehkharghani, Saygin,
  Yanikoglu, and Oflazer}]{2015_Dehkharghani_SentiTurkNet}
Dehkharghani, Rahim, Yucel Saygin, Berrin Yanikoglu, and Kemal Oflazer.
  2016{\natexlab{a}}.
\newblock Sentiturknet: a turkish polarity lexicon for sentiment analysis.
\newblock \emph{Language Resources and Evaluation}, 50(3):667--685.

\bibitem[{Dehkharghani et~al.(2016{\natexlab{b}})Dehkharghani, Yanikoglu,
  Saygin, and Oflazer}]{2016_Dehkharghani_sentiment}
Dehkharghani, Rahim, Berrin Yanikoglu, Yucel Saygin, and Kemal Oflazer.
  2016{\natexlab{b}}.
\newblock Sentiment analysis in turkish at different granularity levels.
\newblock \emph{Natural Language Engineering}, pages 1--25.

\bibitem[{Demir and {\"O}zg{\"u}r(2014)}]{demir2014improving}
Demir, Hakan and Arzucan {\"O}zg{\"u}r. 2014.
\newblock Improving named entity recognition for morphologically rich languages
  using word embeddings.
\newblock In \emph{Machine Learning and Applications (ICMLA), 2014 13th
  International Conference on}, pages 117--122, IEEE.

\bibitem[{Demirtas and Pechenizkiy(2013)}]{Demirtas_2013}
Demirtas, Erkin and Mykola Pechenizkiy. 2013.
\newblock Cross-lingual polarity detection with machine translation.
\newblock In \emph{Proceedings of the Second International Workshop on Issues
  of Sentiment Discovery and Opinion Mining - WISDOM13}, Association for
  Computing Machinery.

\bibitem[{Elman(1990)}]{COGS:COGS203}
Elman, Jeffrey~L. 1990.
\newblock Finding structure in time.
\newblock \emph{Cognitive Science}, 14(2):179--211.

\bibitem[{Erogul(2009)}]{Erogul2009}
Erogul, Umut. 2009.
\newblock Sentiment analysis in turkish.
\newblock Master's thesis, Middle East Technical University.

\bibitem[{Esuli and Sebastiani(2006)}]{Esuli06sentiwordnet:a}
Esuli, Andrea and Fabrizio Sebastiani. 2006.
\newblock Sentiwordnet: A publicly available lexical resource for opinion
  mining.
\newblock In \emph{In Proceedings of the 5th Conference on Language Resources
  and Evaluation (LREC-06}, pages 417--422.

\bibitem[{Firat~Akba and Sever(2014)}]{uccan2014otomatik}
Firat~Akba, Ebru Akcapinar~Sezer, Alaettin~Ucan and Hayri Sever. 2014.
\newblock Assesment of feature selection metrics for sentiment analysis:
  Turkish movie reviews.

\bibitem[{Gers, Schmidhuber, and Cummins(1999)}]{Gers99learningto}
Gers, Felix~A., J{\"u}rgen Schmidhuber, and Fred Cummins. 1999.
\newblock Learning to forget: Continual prediction with lstm.
\newblock \emph{NEURAL COMPUTATION}, 12:2451--2471.

\bibitem[{Giachanou and
  Crestani(2016)}]{2016_Giachanou_Survey_of_Twitter_Sentiment_Analysis}
Giachanou, Anastasia and Fabio Crestani. 2016.
\newblock Like it or not: A survey of twitter sentiment analysis methods.
\newblock \emph{ACM Comput. Surv.}, 49(2):28:1--28:41.

\bibitem[{Hermans and Schrauwen(2013)}]{NIPS2013_5166}
Hermans, Michiel and Benjamin Schrauwen. 2013.
\newblock Training and analysing deep recurrent neural networks.
\newblock In C.~J.~C. Burges, L.~Bottou, M.~Welling, Z.~Ghahramani, and K.~Q.
  Weinberger, editors, \emph{Advances in Neural Information Processing Systems
  26}. Curran Associates, Inc., pages 190--198.

\bibitem[{Hochreiter and Schmidhuber(1997)}]{doi:10.1162/neco.1997.9.8.1735}
Hochreiter, Sepp and J{\"u}rgen Schmidhuber. 1997.
\newblock Long short-term memory.
\newblock \emph{Neural Computation}, 9(8):1735--1780.

\bibitem[{Irsoy and
  Cardie(2014{\natexlab{a}})}]{SC:2014_Irsoy_Deep_Recursive_Neural_Networks_for_Compositionality_in_Language}
Irsoy, Ozan and Claire Cardie. 2014{\natexlab{a}}.
\newblock Deep recursive neural networks for compositionality in language.
\newblock In \emph{Advances in Neural Information Processing Systems 27: Annual
  Conference on Neural Information Processing Systems 2014, December 8-13 2014,
  Montreal, Quebec, Canada}, pages 2096--2104.

\bibitem[{Irsoy and
  Cardie(2014{\natexlab{b}})}]{SC:2014_Irsoy_Opinion_Mining_with_Deep_Recurrent_Neural_Networks}
Irsoy, Ozan and Claire Cardie. 2014{\natexlab{b}}.
\newblock Opinion mining with deep recurrent neural networks.
\newblock In \emph{Proceedings of the 2014 Conference on Empirical Methods in
  Natural Language Processing, {EMNLP} 2014, October 25-29, 2014, Doha, Qatar,
  {A} meeting of SIGDAT, a Special Interest Group of the {ACL}}, pages
  720--728, {ACL}.

\bibitem[{Kalchbrenner, Grefenstette, and
  Blunsom(2014)}]{SC:2014_Kalchbrenner_Convolutional_Neural_Network_for_Modelling_Sentences}
Kalchbrenner, Nal, Edward Grefenstette, and Phil Blunsom. 2014.
\newblock A convolutional neural network for modelling sentences.
\newblock \emph{Proceedings of the 52nd Annual Meeting of the Association for
  Computational Linguistics}.

\bibitem[{Kim(2014)}]{Kim_2014}
Kim, Yoon. 2014.
\newblock Convolutional neural networks for sentence classification.
\newblock In \emph{Proceedings of the 2014 Conference on Empirical Methods in
  Natural Language Processing (EMNLP)}, Association for Computational
  Linguistics (ACL).

\bibitem[{Kisa and Karagoz(2015)}]{onal2015SocialMediaNer}
Kisa, Kezban~Dilek and Pinar Karagoz. 2015.
\newblock Named entity recognition from scratch on social media.
\newblock In \emph{Proceedings of the 6th International Workshop on Mining
  Ubiquitous and Social Environments {(MUSE} 2015) co-located with the 26th
  European Conference on Machine Learning / 19th European Conference on
  Principles and Practice of Knowledge Discovery in Databases {(ECML} {PKDD}
  2015), Porto, Portugal, September 7, 2015.}, pages 2--17.

\bibitem[{Kokciyan et~al.(2013)Kokciyan, Celebi, Ozgur, and
  Uskudarli}]{kokciyan2013bounce}
Kokciyan, Nadin, Arda Celebi, Arzucan Ozgur, and Suzan Uskudarli. 2013.
\newblock Bounce: Sentiment classification in twitter using rich feature sets.
\newblock Citeseer.

\bibitem[{Kuru, Can, and Yuret(2016)}]{2016_kuru_charner}
Kuru, Onur, Ozan~Arkan Can, and Deniz Yuret. 2016.
\newblock Charner: Character-level named entity recognition.
\newblock In \emph{{COLING} 2016, 26th International Conference on
  Computational Linguistics, Proceedings of the Conference: Technical Papers,
  December 11-16, 2016, Osaka, Japan}, pages 911--921.

\bibitem[{Lebret and
  Collobert(2014)}]{2014_Lebret_N_gram_Based_Low_Dimensional_Representation_for_Document_Classification}
Lebret, R{\'{e}}mi and Ronan Collobert. 2014.
\newblock N-gram-based low-dimensional representation for document
  classification.
\newblock \emph{CoRR}, abs/1412.6277.

\bibitem[{Lee, Cho, and Hofmann(2016)}]{DBLP:journals/corr/LeeCH16}
Lee, Jason, Kyunghyun Cho, and Thomas Hofmann. 2016.
\newblock Fully character-level neural machine translation without explicit
  segmentation.
\newblock \emph{CoRR}, abs/1610.03017.

\bibitem[{Li, Jurafsky, and
  Hovy(2015)}]{SC:2015_Li_When_Are_Tree_Structures_Necessary_for_Deep_Learning_of_Representations}
Li, Jiwei, Dan Jurafsky, and Eduard~H. Hovy. 2015.
\newblock When are tree structures necessary for deep learning of
  representations?
\newblock \emph{CoRR}, abs/1503.00185.

\bibitem[{Miller(1995)}]{Miller:1995:WLD:219717.219748}
Miller, George~A. 1995.
\newblock Wordnet: A lexical database for english.
\newblock \emph{Commun. ACM}, 38(11):39--41.

\bibitem[{Schuster and Paliwal(1997)}]{Schuster:1997:BRN:2198065.2205129}
Schuster, M. and K.K. Paliwal. 1997.
\newblock Bidirectional recurrent neural networks.
\newblock \emph{Trans. Sig. Proc.}, 45(11):2673--2681.

\bibitem[{Sennrich, Haddow, and Birch(2015)}]{DBLP:journals/corr/SennrichHB15}
Sennrich, Rico, Barry Haddow, and Alexandra Birch. 2015.
\newblock Neural machine translation of rare words with subword units.
\newblock \emph{CoRR}, abs/1508.07909.

\bibitem[{Socher et~al.(2012)Socher, Huval, Manning, and
  Ng}]{SC:2012_Socher_Semantic_Compositionality_Through_Recursive_Matrix_Vector_Spaces}
Socher, Richard, Brody Huval, Christopher~D Manning, and Andrew~Y Ng. 2012.
\newblock Semantic compositionality through recursive matrix-vector spaces.
\newblock In \emph{Proceedings of the 2012 Joint Conference on Empirical
  Methods in Natural Language Processing and Computational Natural Language
  Learning}, pages 1201--1211, Association for Computational Linguistics.

\bibitem[{Socher et~al.(2011)Socher, Lin, Ng, and Manning}]{SocherEtAl2011:RNN}
Socher, Richard, Cliff~C. Lin, Andrew~Y. Ng, and Christopher~D. Manning. 2011.
\newblock {Parsing Natural Scenes and Natural Language with Recursive Neural
  Networks}.
\newblock In \emph{Proceedings of the 26th International Conference on Machine
  Learning (ICML)}.

\bibitem[{Socher et~al.(2013)Socher, Perelygin, Wu, Chuang, Manning, Ng, and
  Potts}]{SC:2013_Socher_Recursive_Deep_Models_for_Semantic_Compositionality_Over_a_Sentiment_Treebank}
Socher, Richard, Alex Perelygin, Jean Wu, Jason Chuang, Christopher~D. Manning,
  Andrew Ng, and Christopher Potts. 2013.
\newblock Recursive deep models for semantic compositionality over a sentiment
  treebank.
\newblock In \emph{Proceedings of the 2013 Conference on Empirical Methods in
  Natural Language Processing}, pages 1631--1642, Association for Computational
  Linguistics, Seattle, Washington, USA.

\bibitem[{Thelwall et~al.(2010)Thelwall, Buckley, Paltoglou, Cai, and
  Kappas}]{ASI:ASI21416}
Thelwall, Mike, Kevan Buckley, Georgios Paltoglou, Di~Cai, and Arvid Kappas.
  2010.
\newblock Sentiment strength detection in short informal text.
\newblock \emph{Journal of the American Society for Information Science and
  Technology}, 61(12):2544--2558.

\bibitem[{Turney and Pantel(2010)}]{DBLP:journals/corr/abs-1003-1141}
Turney, Peter~D. and Patrick Pantel. 2010.
\newblock From frequency to meaning: Vector space models of semantics.
\newblock \emph{CoRR}, abs/1003.1141.

\bibitem[{Vural et~al.(2012)Vural, Cambazoglu, Senkul, and Tokgoz}]{Vural2012}
Vural, A.~Gural, B.~Barla Cambazoglu, Pinar Senkul, and Z.~Ozge Tokgoz. 2012.
\newblock A framework for sentiment analysis in turkish: Application to
  polarity detection of movie reviews in turkish.
\newblock In \emph{Computer and Information Sciences III}. Springer Nature,
  pages 437--445.

\bibitem[{Wang, Voigt, and Manning(2014)}]{Wang_2014}
Wang, Mengqiu, Rob Voigt, and Christopher~D. Manning. 2014.
\newblock Two knives cut better than one: Chinese word segmentation with dual
  decomposition.
\newblock In \emph{Proceedings of the 52nd Annual Meeting of the Association
  for Computational Linguistics (Volume 2: Short Papers)}, Association for
  Computational Linguistics ({ACL}).

\bibitem[{Yildirimm et~al.(2015)Yildirimm, Cetin, Eryigit, and
  Temel}]{yildirim2015impact}
Yildirimm, Ezgi, Fatih~Samet Cetin, Gulsen Eryigit, and Tanel Temel. 2015.
\newblock The impact of nlp on turkish sentiment analysis.
\newblock \emph{Turkiye Bilisim Vakfi Bilgisayar Bilimleri ve Muhendisligi
  Dergisi}, 7(1 (Basili 8).

\bibitem[{Yin et~al.(2017)Yin, Kann, Yu, and
  Sch{\"{u}}tze}]{DBLP:journals/corr/0001KYS17}
Yin, Wenpeng, Katharina Kann, Mo~Yu, and Hinrich Sch{\"{u}}tze. 2017.
\newblock Comparative study of {CNN} and {RNN} for natural language processing.
\newblock \emph{CoRR}, abs/1702.01923.

\bibitem[{Zhang and Wallace(2015)}]{DBLP:journals/corr/ZhangW15b}
Zhang, Ye and Byron~C. Wallace. 2015.
\newblock A sensitivity analysis of (and practitioners' guide to) convolutional
  neural networks for sentence classification.
\newblock \emph{CoRR}, abs/1510.03820.

\end{thebibliography}
\end{document}